\documentclass[10pt,twocolumn,letterpaper]{article}

\usepackage{wacv}              
\usepackage{tabularx}
\usepackage{lipsum}
\usepackage{verbatim}

\definecolor{wacvblue}{rgb}{0.21,0.49,0.74}
\usepackage[pagebackref,breaklinks,colorlinks,allcolors=wacvblue]{hyperref}
\usepackage{multirow}
\usepackage{makecell}
\usepackage{booktabs}
\usepackage[accsupp]{axessibility} 
\def\wacvPaperID{1768} 
\def\confName{WACV}
\def\confYear{2027}

\title{Recursive CLIP: Vision-Language Distillation with Looped Transformer Blocks}

\author{Szymon Mazurek\textsuperscript{1,2,3}\thanks{Work done during a PhD internship at Samsung AI Center Warsaw.}, Alberto Presta\textsuperscript{1}, Michal{} Byra\textsuperscript{1,4} \\
\textsuperscript{1} Samsung AI Center Warsaw, Poland \quad
\textsuperscript{2} AGH University of Krakow, Poland \\
\textsuperscript{3} ACC Cyfronet AGH, Poland \\
\textsuperscript{4} Institute of Fundamental Technological Research, Polish Academy of Sciences, Poland \\
{\tt\small szmazurek@agh.edu.pl, a.presta@samsung.com, m.byra@samsung.com}
}

\begin{document}
\maketitle
\begin{abstract}

Contrastive vision--language models such as CLIP typically rely on independently parameterized transformer blocks across depth. We investigate the extent to which these blocks can instead be reused recurrently while preserving the transferable capabilities of CLIP. We study recursive transformer backbones in which a small set of $N$ transformer blocks is applied $M$ times per forward pass. By holding $N \cdot M$ fixed to the depth of a standard ViT-B, our models preserve computational depth while progressively increasing parameter sharing. We consider two recursive CLIP model families: \emph{rCLIP}, where only the vision encoder is recursive, and \emph{r²CLIP}, where recurrence is also introduced in the text encoder. We study these models under both direct contrastive training and through knowledge distillation from a strong ViT-B/16 teacher pretrained on DFN-2B, using the substantially smaller CC3M and CC12M datasets. We evaluate the resulting models on zero-shot classification and cross-modal retrieval, and further study their adaptability through few-shot learning, visual grounding, and concept unlearning. Our results show that CLIP can tolerate substantial recurrent block reuse, although its effect depends on the encoder and downstream task. Recurrence also provides parameter efficiency by construction, since fewer distinct transformer blocks need to be stored and adapted. Our results establish recurrent block reuse as a practical design space for vision--language models and highlight its implications for knowledge transfer, downstream adaptation, and parameter efficiency.

\end{abstract}

\begin{figure}[h]
    \centering
    \includegraphics[width=0.75\linewidth]{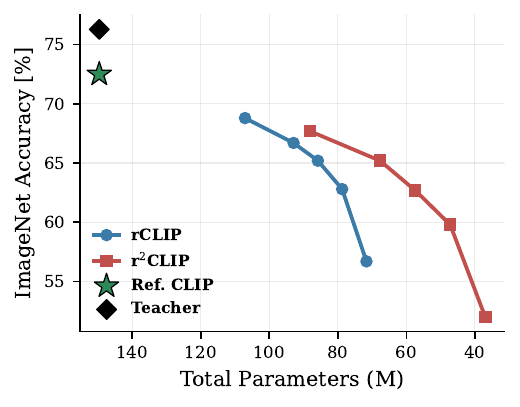}
\caption{Accuracy--size trade-off of recursive CLIP models. Zero-shot ImageNet top-1 accuracy versus total parameter count for rCLIP and r²CLIP, compared with the DFN-2B ViT-B/16 teacher and distilled Reference CLIP. The parameter axis is reversed, with smaller models shown to the right.}    
    \label{fig:pareto-front}
\end{figure}

\section{Introduction}
\label{sec:intro}

Vision--language models have become a central paradigm for learning transferable representations from paired visual and textual data. Among them, CLIP~\cite{radford2021learning} has demonstrated strong transfer across a wide range of downstream tasks. Its vision and text encoders typically rely on transformer architectures~\cite{vaswani2017attention} composed of multiple independently parameterized blocks. While this design provides substantial representational capacity, it remains unclear to what extent distinct parameters are required at every depth to preserve the transferable capabilities of CLIP.

Recursive, or looped, transformers provide an alternative by repeatedly applying a smaller set of shared blocks across depth \cite{dehghani2018universal}. In this work, we investigate the extent to which recurrence can be introduced into CLIP while preserving its vision--language capabilities. This provides a controlled way to study the effect of depth-wise parameter sharing on representation quality, knowledge transfer, and downstream adaptation. This parameter sharing also improves parameter efficiency, as increasing recurrence reduces the number of unique transformer blocks in the model.

We construct recursive CLIP models by replacing the $L$ transformer blocks of an encoder with $N$ distinct blocks executed $M$ times, such that $N \cdot M = L$. We consider two model families: \textbf{rCLIP}, where only the vision encoder is recursive, and \textbf{r²CLIP}, where both the vision and text encoders are recursive. By varying $N$, we study different degrees of recurrence while keeping the computational depth of the corresponding ViT-B architecture fixed. We study these models both under direct contrastive training and through knowledge distillation from a strong DFN-2B ViT-B/16 teacher, allowing us to investigate how effectively representations learned by a conventional CLIP model can be transferred to recursive students.

Beyond standard zero-shot classification and cross-modal retrieval, we investigate how recurrence affects downstream adaptation through few-shot learning, visual grounding, and concept unlearning. We also examine whether distillation objectives developed for conventional transformer students remain suitable when the same blocks are reused across depth. The parameter efficiency resulting from the recurrence is illustrated in Fig.~\ref{fig:pareto-front}, which relates model size to zero-shot ImageNet performance for rCLIP and r²CLIP. Block reuse also extends naturally to parameter-efficient adaptation, since adapters attached to shared blocks are reused across recurrent executions.

Our contributions are as follows:

\begin{itemize}

\item We systematically study recursive CLIP architectures based on depth-wise block reuse, considering both vision-only and joint vision--text variants, and investigate how extensively recurrence can be introduced into CLIP while preserving its vision--language capabilities.

\item We investigate distillation from conventional CLIP into recursive students, analyzing how recurrence and vision/text sharing affect knowledge transfer. 

\item We evaluate recursive and non-recursive distilled CLIP models beyond standard zero-shot classification and retrieval, extending the analysis to few-shot learning, visual grounding, and concept unlearning.

\item We highlight the parameter and adaptation efficiency that follows naturally from recurrent block reuse, reducing both stored backbone parameters and the number of distinct adapters required by methods such as LoRA.

\end{itemize}

\begin{figure*}
    \centering
    \includegraphics[width=0.65\linewidth]{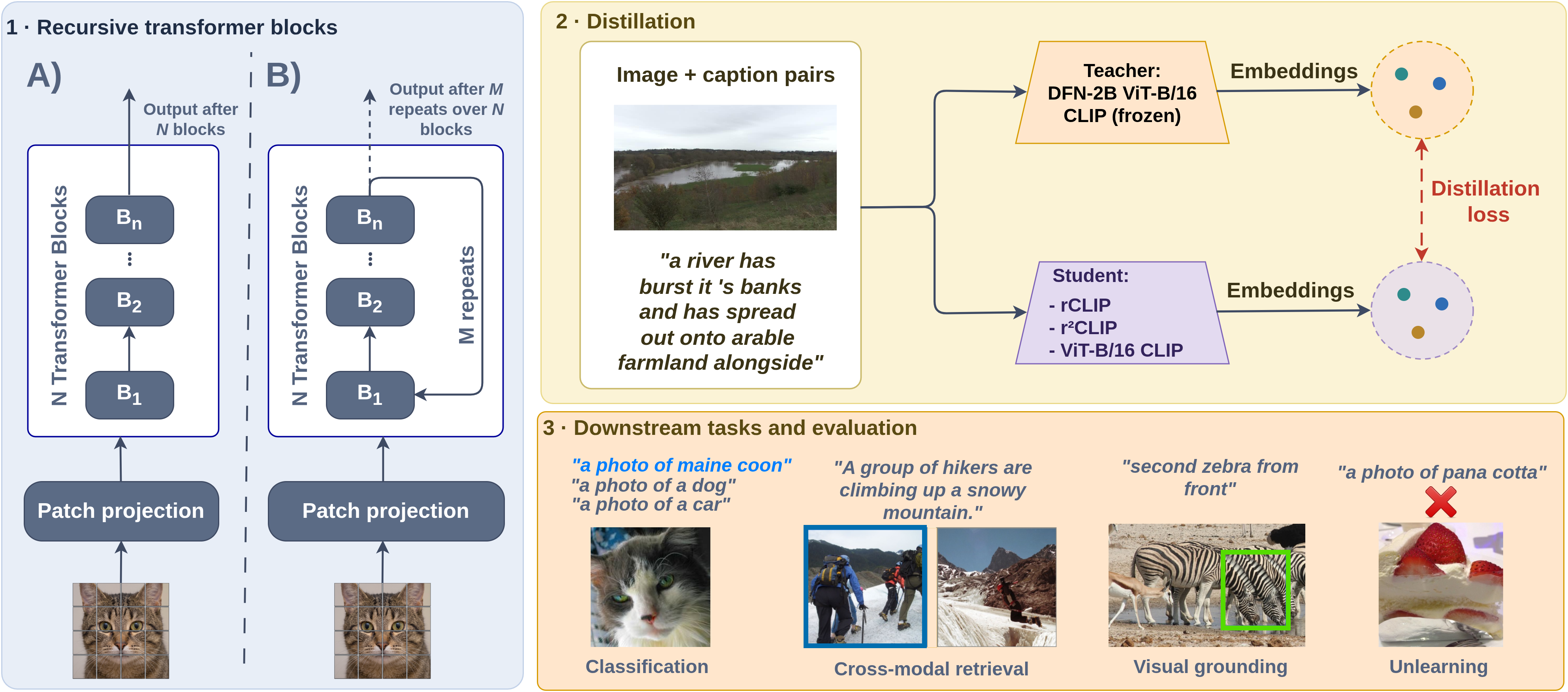}
    \caption{Overview of the experimental approach. \textbf{(1)} A recursive encoder replaces L distinct transformer blocks (\textbf{A}) with N blocks applied M times, N·M = L (\textbf{B}), keeping computational depth fixed while reducing stored parameters. \textbf{(2)} Recursive (rCLIP, r²CLIP) and non-recursive students are distilled from a frozen DFN-2B ViT-B/16 teacher on CC3M + CC12M. \textbf{(3)} The resulting models are evaluated on classification (zero- and few-shot adapted), cross-modal retrieval, visual grounding and concept unlearning.}
    \label{fig:overview}
\end{figure*}
\section{Related work}
\label{sec:related}

\noindent\textbf{\emph{Contrastive language--image pre-training and compression.}}
CLIP introduced large-scale contrastive language--image pre-training, learning aligned visual and textual representations from paired image--caption data and enabling strong zero-shot transfer across visual tasks~\cite{radford2021learning}. Subsequent works have improved the scale, data quality, and training of contrastive vision--language models~\cite{jia2021scaling,fang2024data,zhai2023sigmoid}. Their increasing model and data requirements have also motivated methods for constructing smaller and more efficient CLIP models. TinyCLIP combines cross-modal affinity mimicking with weight inheritance and progressive distillation~\cite{wu2023tinyclip}, while CLIP-KD systematically studies different forms of knowledge transfer between CLIP teachers and students~\cite{yang2024clip}. Other work reduces inference cost through efficient hybrid backbones~\cite{vasu2024mobileclip} or by token pruning~\cite{wu2025patch}. In contrast, we preserve the effective computational depth of ViT and progressively replace depth-specific transformer blocks with recurrent applications of shared blocks. This allows us to study the extent to which recurrence can be introduced into CLIP, including how conventional CLIP representations can be transferred to recursive students through distillation and how the resulting models adapt to downstream tasks. Parameter reduction follows directly from the reuse of the same blocks across depth.

\noindent\textbf{\emph{Recursive transformers and parameter sharing.}}
Recurrence provides an alternative to conventional depth by repeatedly applying shared transformations. The Universal Transformer applies transformer layers recurrently across computational steps~\cite{dehghani2018universal}, while ALBERT demonstrates extensive cross-layer parameter sharing in language models~\cite{Lan2020ALBERT},
which recent work relaxes via loop-specific LoRA adapters~\cite{bae2025relaxed}.
More recent looped transformers study repeated latent computation in reasoning, algorithmic tasks and in-context learning~\cite{giannou2023looped,yang2023looped,saunshi2025reasoning} and in scaling effective depth at inference time~\cite{geiping2025scaling}.

Recurrence and weight sharing have also been explored in vision transformers (ViTs). SReT introduces recursive weight sharing across depth and combines it with sliced group self-attention and non-shared nonlinear projection layers to improve parameter utilization while controlling the computational cost of recurrence~\cite{shen2022sliced}. Its main architecture therefore does not simply repeat an unchanged standard ViT block, although naive recursion is also studied. MiniViT similarly shares transformer parameters while introducing lightweight layer-specific transformations to preserve representation diversity~\cite{zhang2022minivit}. More restrictive forms of recurrence are studied in bViT, where standard transformer blocks are directly reused across the effective ViT depth without additional layer-specific transformations~\cite{byra2026bvit}, and in Raptor, which constructs recurrent approximations of pretrained DINO vision through distillation~\cite{jacobs2026block}. Together, these works show that recurrent computation can substantially reduce layer-specific parameterization in ViTs.

Weight sharing has received less attention in vision--language models. Gao et al.~\cite{gao2023learning} study cross-layer and cross-modal weight sharing together with pruning in grounding-oriented vision--language transformers, while R-MMA~\cite{fahim2026rmma} shares recurrent multimodal adapters across layers of frozen CLIP image and text encoders without making the CLIP backbone recurrent. In contrast, we study recurrence directly in the CLIP encoders. We vary the number of distinct blocks and recurrent executions at fixed computational depth, consider both vision-only rCLIP and joint vision--text r²CLIP architectures, and investigate their behavior under knowledge distillation and across downstream tasks.

\section{Methods}
\label{sec:method}

As shown in Fig.~\ref{fig:overview}, we study recursive CLIP architectures and their downstream use. Sect.~\ref{method:3.1} introduces the vision-only and vision--text recursive variants, Sect.~\ref{method:3.2} describes the teacher--student distillation framework, and Sect.~\ref{method:3.3} presents the downstream evaluation setup.

\subsection{Recursive transformers} \label{method:3.1}




Following prior works on parameter sharing in depth~\cite{shen2022sliced,jacobs2026block,byra2026bvit}, we develop recursive CLIP encoders by replacing the sequence of distinct transformer blocks with a smaller set of blocks that is repeatedly applied. The input tokenization and output layer are kept identical to those of the standard CLIP architecture. We define the recursive transformer core as a sequence of $N$ distinct blocks: 

\begin{equation} 
F = B_N \circ B_{N-1} \circ \cdots \circ B_1, 
\end{equation} 
whose parameters are shared across $M$ recurrent applications. Given the initial token representation: 
\begin{equation} 
A^{0} = \mathcal{T}(x), 
\end{equation} 

\noindent where $\mathcal{T}$ denotes the tokenizer for input $x$, the recursive computation is given by: 

\begin{equation} 
A^{m} = F(A^{m-1}), \qquad m=1,\ldots,M.
\label{eq:recursion}
\end{equation}

We point out that, in contrast to previous works, we also consider multiple distinct blocks applied recursively, i.e., $N>1$; this allows us to explore the trade-off between performance and model memory footprint. We highlight that no iteration-specific embeddings are introduced, so successive applications of $F$ differ only through their input activations.
The final representation is obtained as $z=\mathcal{P}(A^M)$, where $\mathcal{P}$ denotes the standard CLIP output mapping. We choose $N$ and $M$ such that $N\cdot M=L$, where $L$ is the number of transformer blocks in the corresponding reference encoder. For the ViT-B/16 CLIP architecture, both the vision and text encoders contain $L=12$ blocks. We therefore consider $(N,M)\in\{(1,12),(2,6),(3,4),(4,3),(6,2)\}$. Since all configurations execute 12 transformer blocks per forward pass, they are FLOP-matched to the non-recursive reference while requiring fewer distinct transformer parameters. This reduction in model size is particularly relevant for deployment on memory-constrained edge devices, where storing large vision--language models can be prohibitive. Table~\ref{tab:model-sizes} reports the resulting parameter counts. We denote each configuration as \emph{$N$b$M$i}, where $N$ is the number of distinct blocks and $M$ the number of iterations. For example, \emph{1b12i} denotes one block applied 12 times, while \emph{6b2i} denotes six blocks applied twice.

\begin{table}[htbp]
\centering
\caption{Parameter counts across rCLIP and r²CLIP configurations. Each model uses 17.57 visual and 2.98 text GMACs at $224\times224$ resolution and 77-token context.}
\small
\setlength{\tabcolsep}{5pt}
\renewcommand{\arraystretch}{1.0}
\begin{tabular}{lccc}
\hline
CLIP Model & Visual [M] & Text [M] & Total [M] \\
\hline
Ref. CLIP    & 86.19 & 63.43 & 149.62 \\
\hline
rCLIP 1b12i  & 8.22  & 63.43 & 71.65 \\
rCLIP 2b6i   & 15.31 & 63.43 & 78.74 \\
rCLIP 3b4i   & 22.40 & 63.43 & 85.83 \\
rCLIP 4b3i   & 29.48 & 63.43 & 92.91 \\
rCLIP 6b2i   & 43.65 & 63.43 & 107.08 \\
\hline
r²CLIP 1b12i & 8.22  & 28.75 & 36.97 \\
r²CLIP 2b6i  & 15.31 & 31.90 & 47.21 \\
r²CLIP 3b4i  & 22.40 & 35.05 & 57.45 \\
r²CLIP 4b3i  & 29.48 & 38.20 & 67.68 \\
r²CLIP 6b2i  & 43.65 & 44.51 & 88.16 \\
\hline
\end{tabular}
\label{tab:model-sizes}
\end{table}

\subsection{Model training and knowledge distillation} \label{method:3.2}

We rely on knowledge distillation \cite{hinton2015distilling} (KD), adapting the CLIP-KD framework \cite{yang2024clip} to distill a CLIP ViT-B/16 trained on DFN-2B  (\textbf{Teacher}) \cite{fang2024data} into two recursive CLIP families, replacing either the vision encoder alone (\textbf{{rCLIP}}) or both encoders (\textbf{{r²CLIP}}), on the combined CC3M \cite{sharma2018conceptual} and CC12M \cite{changpinyo2021conceptual} datasets, totaling 13,874,493 image-caption pairs. As a reference, we also distill into a randomly initialized ViT-B/16 CLIP model (\textbf{{Reference CLIP}}).

Following CLIP-KD~\cite{yang2024clip}, we consider several forms of teacher
supervision in addition to the standard CLIP contrastive objective. Feature
Distillation (FD) directly matches the final image and text embeddings of the
teacher and student. Contrastive Relational Distillation (CRD) instead matches
their cross-modal similarity distributions over the mini-batch. Interactive
Contrastive Learning (ICL) contrasts student image features against teacher
text features and, symmetrically, student text features against teacher image
features. We additionally evaluate Activation Matching
(AM)~\cite{romero2015fitnets}, which aligns intermediate vision
representations with those of the teacher. The general objective is as follows:
\begin{equation}
\begin{aligned}
\mathcal{L}
&=
\mathcal{L}_{\mathrm{CLIP}}
+ \lambda_{\mathrm{CRD}}\mathcal{L}_{\mathrm{CRD}}
+ \lambda_{\mathrm{ICL}}\mathcal{L}_{\mathrm{ICL}} \\
&\quad
+ \lambda_{\mathrm{FD}}\mathcal{L}_{\mathrm{FD}}
+ \alpha\mathcal{L}_{\mathrm{AM}} .
\end{aligned}
\label{eq:kd-objective}
\end{equation}
Here, $\mathcal{L}_{\mathrm{CLIP}}$ is the standard symmetric image--text
contrastive loss. Full definitions of all terms are provided in the
Supplementary Material. We select the distillation terms and their weights in
Sect.~\ref{sec:loss-choice}.

AM also requires a correspondence between intermediate student
executions and teacher layers. For block $B_i$ in recursive iteration $m$, we
assign the global execution index
\begin{equation}
\ell=(m-1)N+i,
\qquad \ell=1,\ldots,NM=L .
\label{eq:unroll}
\end{equation}
This gives an index-preserving correspondence between the $L$ student
executions and the $L$ teacher layers. Unlike in a conventional transformer,
a recursive block $B_i$ is matched to multiple teacher layers,
$\{i,N+i,\ldots,(M-1)N+i\}$. This alignment therefore assumes that successive
applications of a shared block traverse representations corresponding to the
teacher's layer-wise trajectory. We examine this assumption empirically in
Sect.~\ref{sec:loss-choice}; the full AM definition is provided in the
Supplementary Material.

We otherwise follow the training setup of~\cite{yang2024clip}, using
AdamW~\cite{loshchilov2018decoupled} with a peak learning rate of $10^{-3}$,
10k warmup steps followed by cosine decay, and an effective batch size of
1024. Models are trained for 32 epochs. Additional training details are
provided in the Supplementary.

\subsection{Downstream tasks} \label{method:3.3}

\textbf{\emph{Zero-shot classification and retrieval.}} We report zero-shot top-1 accuracy on ImageNet-1k~\cite{russakovsky2015imagenet} and its distribution-shifted variants (ImageNet-V2~\cite{recht2019imagenet}, ImageNet-Rendition~\cite{hendrycks2021many}, ImageNet-Sketch~\cite{wang2019learning}), following the standard CLIP protocol with an 80-prompt template ensemble~\cite{radford2021learning,yang2024clip}. We additionally report Recall@1 for text-to-image and image-to-text retrieval on Flickr30k~\cite{young2014image} and MSCOCO~\cite{lin2014mscoco}, using the splits of~\cite{karpathy2015deep}.



\noindent\textbf{\emph{Few-shot learning for classification.}} 
We adapt each backbone with CLIP-LoRA~\cite{zanella2024low} at 16 shots per class on ten standard classification benchmarks, following the original recipe. LoRA adapters are attached to every distinct block of the recursive core and shared across its iterations, exactly as the frozen weights are. Because CLIP-LoRA uses a single prompt template rather than the 80-template ensemble, the zero-shot numbers reported alongside these results are slightly lower than those of Sect.~\ref{sec:distillation}; full adaptation details are given in the Supplementary Material.

\noindent\textbf{\emph{Visual grounding.}}
We adapt each backbone to visual grounding using HiVG~\cite{xiao2024hivg} on RefCOCO, RefCOCO+\cite{yu2016modeling}, RefCOCOg\cite{mao2016generation}, ReferItGame~\cite{kazemzadeh2014referitgame}, and Flickr30k Entities~\cite{plummer2015flickr30k}, reporting top-1 accuracy at IoU $\geq 0.5$. Since HiVG's cross-modal bridges and LoRA curriculum are defined with respect to distinct transformer layers, we instead attach them to execution steps and to the $N$ distinct blocks respectively, so recursive backbones remain directly comparable to the non-recursive baseline. Full adaptation details are given in the Supplementary.

\noindent\textbf{\emph{Unlearning.}} We evaluate concept unlearning using DIET~\cite{kunananthaseelan2026diet}, a retain-data-free method that removes a class via LoRA adapters attached to the vision encoder, again one per distinct block. We report accuracy on the forget class ($D_f$) and on the full test set ($D_t$), before and after unlearning, averaged over five seeds with a different randomly drawn class per seed. Full training and adaptation details are given in the Supplementary Material.

\noindent\textbf{\emph{Reference methods.}}
In all downstream adaptation experiments (few-shot classification, grounding and unlearning), we additionally evaluate three non-recursive CLIP models alongside the recursive backbones: the OpenAI ViT-B/16 checkpoint, which is the standard starting point in the methods we adapt, our Teacher model, and the non-recursive Reference CLIP we distilled from the teacher.


\section{Experiments}
\label{sec:experiments}

We organize the evaluation in three stages. We first train the full block/iteration grid directly under a CLIP objective, with no teacher, to characterize the intrinsic behavior of recursive backbones. We then establish the distillation procedure and train the same configurations by distilling a DFN-2B ViT-B/16 teacher. We measure how much of the teacher's transferable representation each configuration recovers. Finally, we evaluate the distilled checkpoints on downstream tasks. We report both the rCLIP and the r²CLIP performance scores for various block/iteration configurations, and compare against a FLOP-matched non-recursive ViT-B/16 CLIP model. For brevity, in the downstream task evaluations we provide the results only for 1b12i, 3b4i, and 6b2i to highlight relevant trends. Extended results for all configurations can be found in the Supplementary. 

\subsection{Baseline CLIP training}
\label{sec:baseline}
\begin{figure}[h]
    \centering
    \includegraphics[width=0.65\linewidth]{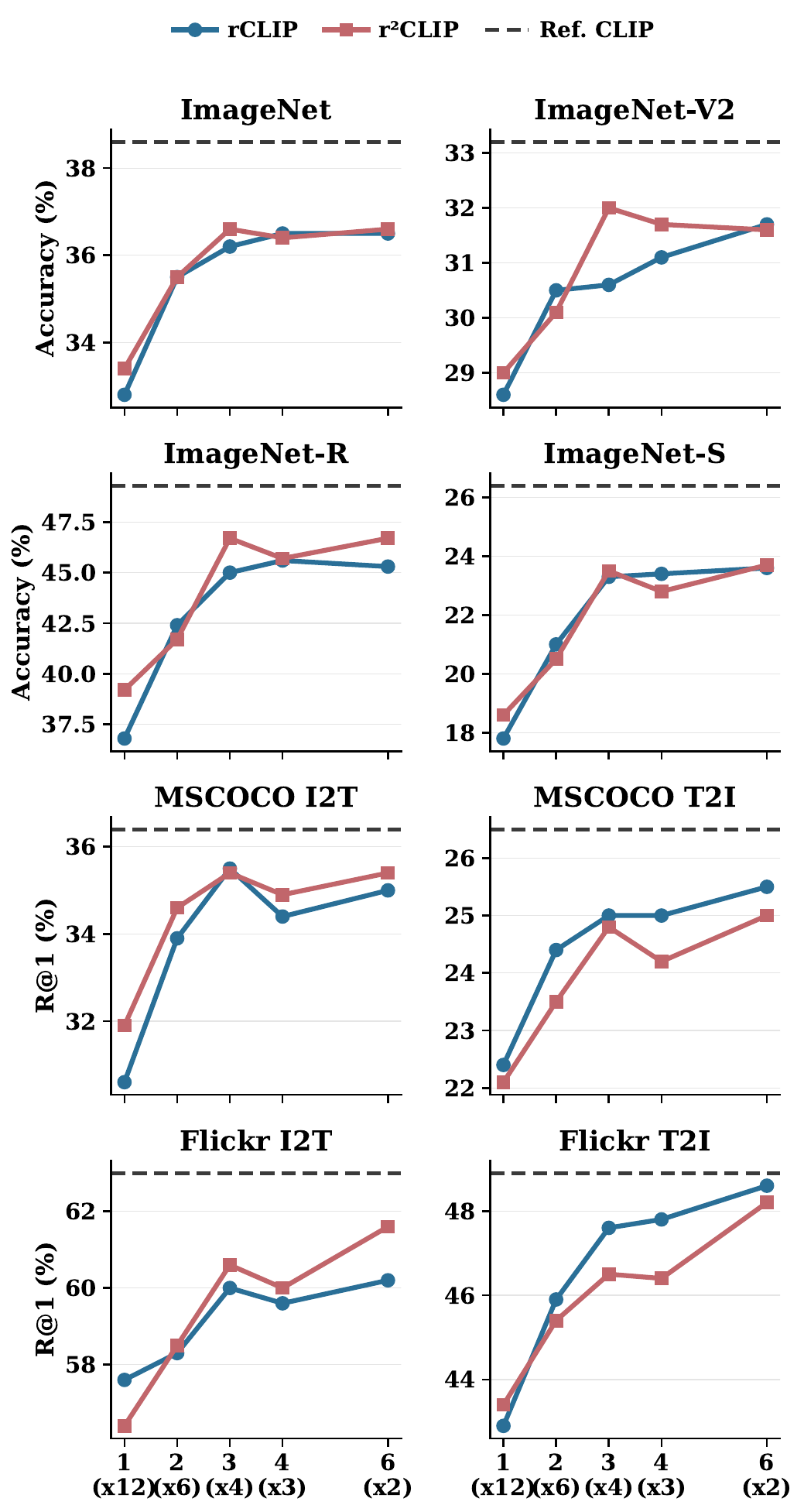}
    \caption{Baseline results for different rCLIP and r²CLIP configurations trained directly with the CLIP objective.}
    \label{fig:baseline-clip}
\end{figure}

We first investigate recursive models trained without distillation using CLIP objective, with the non-recursive ViT-B CLIP as the reference. Fig. \ref{fig:baseline-clip} shows zero-shot accuracy on ImageNet and its variants, with cross-modal retrieval.
Performance consistently increases with the number of unique blocks. For rCLIP, ImageNet accuracy increases from 32.8\% with one block to 36.5\% with six blocks, with the largest gain occurring between one and two blocks. Similar saturation is observed on the shifted variants and retrieval tasks, indicating that most of the gap to ViT-B is recovered once several distinct blocks are introduced.
Recursive text tower causes little additional degradation despite substantially reducing the parameter count. In the single block configuration, r²CLIP even slightly outperforms rCLIP, while from two blocks onward their performance is nearly identical. This suggests that the text encoder can tolerate extensive weight sharing, with little cost once at least two distinct blocks are retained.

\subsection{Choosing the distillation objective}
\label{sec:loss-choice}

We first determine which distillation signals transfer effectively to a recursive student. Each component of Eq.~\eqref{eq:kd-objective} is evaluated individually on top of the CLIP contrastive loss, together with the full CLIP-KD composition of CRD + ICL + FD~\cite{yang2024clip}. We compare the non-recursive Reference CLIP with the most constrained recursive model, rCLIP 1b12i. All runs in this analysis use CC3M, identical optimization settings and a single seed. Table~\ref{tab:loss-composition} reports the results; weighting experiments are provided in the Supplementary. FD is the strongest individual distillation signal for both architectures. For Reference CLIP, FD reaches 37.0\% ImageNet accuracy, nearly matching the 37.3\% obtained with the  CRD + ICL + FD objective. CRD and ICL are weaker when used individually. The same ordering is preserved for rCLIP 1b12i, although all three objectives transfer less effectively to the recursive student.
AM behaves qualitatively differently. On Reference CLIP it is the second strongest individual objective, reaching 31.0\% ImageNet accuracy, whereas on rCLIP 1b12i it falls to 15.5\%, below CRD, ICL and FD. Under AM, the recursive model retains only 50\% of the corresponding non-recursive accuracy, compared with 76--95\% for the output-level objectives. This result suggests that layer-wise activation correspondence is poorly suited to strong parameter sharing. Under Eq.~\eqref{eq:unroll}, the single block of rCLIP 1b12i is required to match twelve distinct teacher layers across its iterations. The poor AM result indicates that the recurrent trajectory does not reproduce the sequence of representations generated by twelve independently parameterized teacher blocks. The pointwise matching of that trajectory therefore acts as an overly restrictive constraint. We leave the detailed explanation of this phenomenon to future research. We consequently exclude AM from the remaining experiments. A subsequent weighting sweep over the output-level objectives selects $\lambda_{\mathrm{CRD}}=1$, $\lambda_{\mathrm{ICL}}=0$, and
$\lambda_{\mathrm{FD}}=10^4$. This configuration is used for all subsequent distillation experiments, with the experimental details found in the Supplementary Material.

\begin{table}[t]
\centering
\caption{Distillation loss components evaluated on the non-recursive Reference CLIP and the recursive rCLIP 1b12i student. Each objective is added to the primary CLIP contrastive loss with weight 1. CRD, ICL and FD use the weights of~\cite{yang2024clip} ($1$, $1$ and $2000$), while AM uses $\alpha=2000$. The top row reports the full CLIP-KD composition. \emph{Ret.} denotes recursive ImageNet accuracy as a fraction of the corresponding non-recursive accuracy.}
\setlength{\tabcolsep}{4pt}
\small
\begin{tabular}{lccccc}
\toprule
& \multicolumn{2}{c}{Ref. CLIP} & \multicolumn{2}{c}{rCLIP 1b12i} & \\
\cmidrule(lr){2-3}\cmidrule(lr){4-5}
Objective & IN & IN-R & IN & IN-R & Ret. \\
\midrule
CRD + ICL + FD & 37.3 & 46.9 & 30.3 & 33.2 & 81\% \\
FD             & 37.0 & 46.3 & 28.2 & 29.2 & 76\% \\
CRD            & 26.2 & 28.8 & 22.4 & 22.6 & 86\% \\
ICL            & 19.3 & 22.3 & 18.4 & 19.6 & 95\% \\
AM             & 31.0 & 47.5 & 15.5 & 17.0 & \textbf{50\%} \\
\bottomrule
\end{tabular}

\label{tab:loss-composition}
\end{table}

\subsection{Distillation results}
\label{sec:distillation}

\begin{figure}[h]
    \centering
    \includegraphics[width=0.65\linewidth]{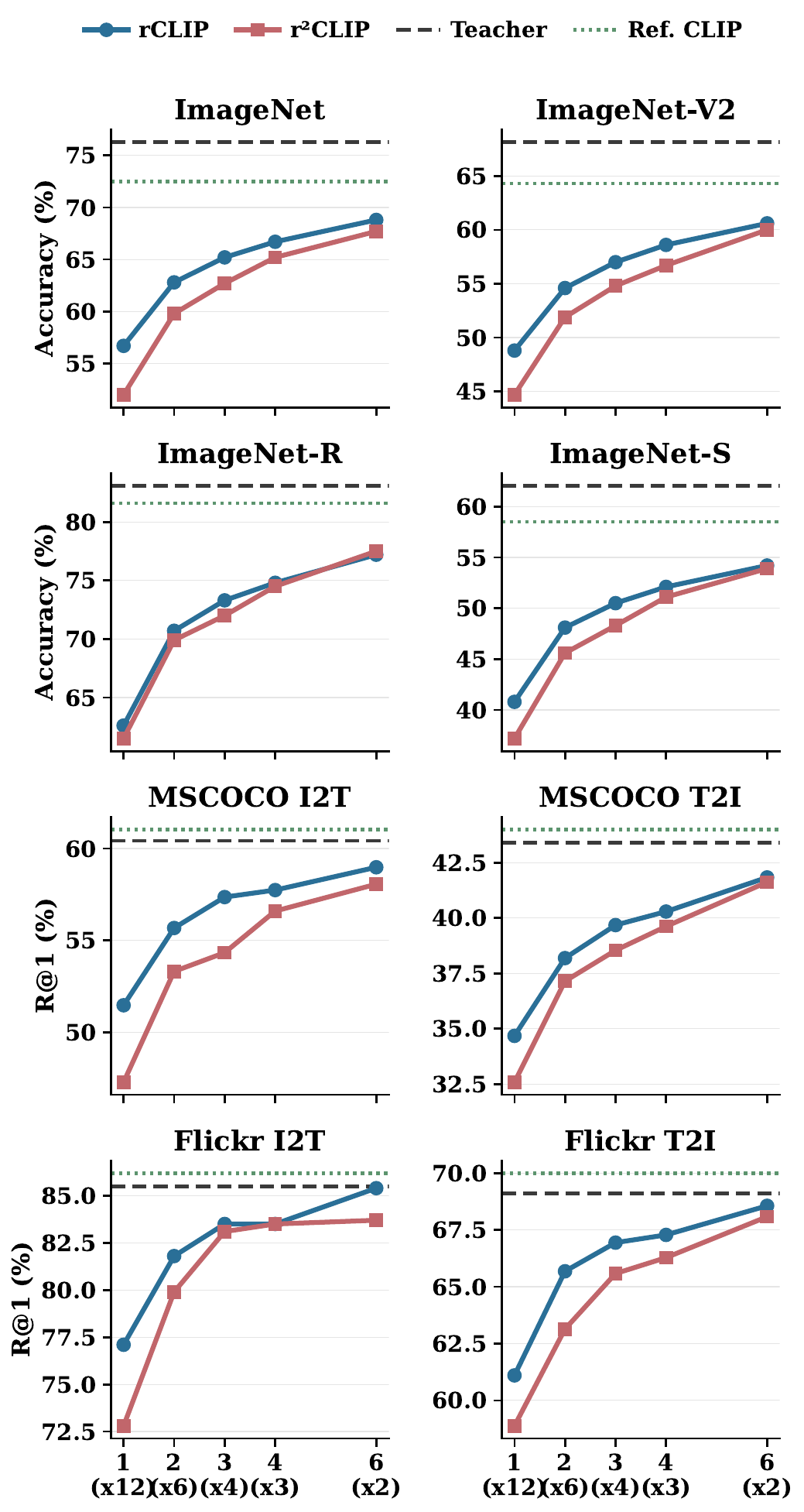}
    \caption{Results for the different rCLIP and r²CLIP configurations distilled from the DFN-2B ViT-B/16 teacher.}
    \label{fig:distillation-results}
\end{figure}

We next train the same architectures with distillation from the DFN-2B ViT-B/16 teacher, using the loss selected in Sect.~\ref{sec:loss-choice} and Sup. Material. Fig.~\ref{fig:distillation-results} reports the same eight metrics as Sect. \ref{sec:baseline}, together with the teacher and a non-recursive ViT-B reference distilled using the same recipe. Distillation improves all models, while revealing a larger gap between recursive and non-recursive students. With one block, rCLIP and r²CLIP reach 56.7\% and 52.0\% ImageNet zero-shot accuracy, compared with 72.5\% for the distilled ViT-B. Performance increases with the number of unique blocks, with rCLIP reaching 68.8\% at six blocks and r²CLIP following the same trend. Unlike in CLIP-only training, rCLIP retains an advantage over r²CLIP, although the ImageNet gap decreases from 4.7 points at one block to 1.1 at six. The difference is even smaller on the shifted datasets, with r²CLIP slightly outperforming rCLIP on ImageNet-R at six blocks. A larger gap remains for retrieval: Flickr I2T reaches 85.4 for rCLIP and 83.7 for r²CLIP, suggesting that aggressive weight sharing in the text tower affects cross-modal matching more strongly. The distillation recipe is also effective for the non-recursive student, which remains at most 4 percentage points below the teacher in zero-shot classification despite substantially less training data, while slightly surpassing it on cross-modal retrieval.

\subsection{Downstream task adaptation} 
\label{sec:downstream-tasks}

\textbf{\emph{Few-shot learning for classification.}} Table \ref{tab:fewshot-results} reports zero-shot and 16-shot adapted accuracy across ten datasets for the recursive and reference models. In the zero-shot setting, performance increases with the number of blocks. For rCLIP, ImageNet accuracy improves from 56.34\% at one block to 68.54\% at six blocks, approaching the Reference CLIP at 71.86\%, with mostly the same monotone trend across the remaining datasets. r²CLIP remains a few points below the corresponding rCLIP configuration on most datasets, although the gap narrows with more blocks, indicating that text-tower recurrence is most costly under strong sharing. The gap to the distilled reference is largest on fine-grained datasets: rCLIP 6b2i retains 93\% of the reference accuracy on Food101 (75.39 vs 81.05) and 78\% on Aircraft (12.87 vs 16.59), compared with 98\% on Caltech101 and Pets. Recursion therefore affects fine-grained recognition more strongly, while coarser classification is largely preserved. Few-shot adaptation with CLIP-LoRA substantially closes the gap to the distilled ViT-B CLIP. At sixteen shots, rCLIP 6b2i reaches 72.13 on ImageNet versus 75.18 for the reference, and comes within one to two points on several datasets, including Flowers (98.09 vs 99.03) and Caltech101 (96.63 vs 97.24). rCLIP clearly outperforms r²CLIP in the 1b12i configuration, but this gap becomes small as more blocks are retained. Overall, the recovery under few-shot adaptation indicates that recursive backbones retain adaptable representations despite their reduced parameter count.

\begin{table*}[h!]
\centering
\caption{Zero-shot and 16-shot accuracy [\%] across ten classification datasets.
Few-shot results use CLIP-LoRA~\cite{zanella2024low} adaptation method. Teacher denotes the DFN-2B ViT-B/16,
Ref.\ CLIP the non-recursive ViT-B/16 distilled from it under the same recipe as the
recursive students. Within each setting, \textbf{bold} marks the best recursive model per
column and \underline{underline} the best model overall. The remaining block/iteration
configurations are reported in the Supplementary.}
\label{tab:fewshot-results}
\resizebox{\textwidth}{!}{%
\begin{tabular}{llcccccccccc}
\toprule
Setting & CLIP Model & ImageNet & SUN & Aircraft & EuroSAT & Food
& Pets & Flowers & Caltech & DTD & UCF \\
\midrule
\multirow{9}{*}{Zero-shot}
& ViT-B/16 (OAI chkpt) & 66.73 & 62.57 & \underline{23.97} & 42.09 & 85.27 & 88.28 & 67.28 & 93.23 & 44.39 & \underline{65.16} \\
& Teacher & \underline{75.91} & \underline{70.68} & 21.60 & \underline{48.09} & \underline{88.27} & 91.47 & \underline{79.09} & \underline{96.96} & \underline{56.86} & 61.83 \\
& Ref. CLIP & 71.86 & 70.45 & 16.59 & 44.43 & 81.05 & \underline{91.66} & 66.10 & 95.46 & 55.67 & 60.19 \\
\cmidrule(lr){2-12}
& rCLIP 1b12i  & 56.34 & 61.22 & 4.35  & 33.53 & 58.80 & 84.57 & 48.76 & 89.45 & 42.49 & 49.38 \\
& rCLIP 3b4i   & 64.65 & 66.30 & 8.73  & \textbf{41.88} & 70.63 & 87.24 & 59.11 & 92.54 & 50.77 & 53.61 \\
& rCLIP 6b2i   & \textbf{68.54} & 68.22 & \textbf{12.87} & 39.83 & \textbf{75.39} & \textbf{89.81} & \textbf{61.92} & 93.55 & \textbf{54.43} & 56.54 \\
\cmidrule(lr){2-12}
& r²CLIP 1b12i & 51.05 & 59.11 & 4.62  & 32.01 & 53.64 & 80.62 & 35.57 & 88.24 & 37.65 & 47.32 \\
& r²CLIP 3b4i  & 62.05 & 65.01 & 8.97  & 33.99 & 68.41 & 87.33 & 54.61 & 92.21 & 46.04 & 53.32 \\
& r²CLIP 6b2i  & 66.93 & \textbf{68.40} & 12.66 & 36.81 & 75.10 & 86.13 & 59.64 & \textbf{94.04} & 51.48 & \textbf{56.89} \\
\midrule\midrule
\multirow{9}{*}{Few-shot (16)}
& ViT-B/16 (OAI chkpt) & 73.37 & 76.22 & 55.54 & 90.65 & 84.74 & 91.96 & 98.01 & 96.39 & 71.99 & \underline{86.23} \\
& Teacher & \underline{77.83} & 76.92 & \underline{61.60} & \underline{91.58} & \underline{87.11} & 92.53 & \underline{99.27} & \underline{97.53} & 75.83 & 83.11 \\
& Ref. CLIP & 75.18 & \underline{77.21} & 58.78 & 90.74 & 84.58 & \underline{92.78} & 99.03 & 97.24 & \underline{75.95} & 82.42 \\
\cmidrule(lr){2-12}
& rCLIP 1b12i  & 61.45 & 69.43 & 35.13 & 87.09 & 67.61 & 85.47 & 94.07 & 94.08 & 61.29 & 72.98 \\
& rCLIP 3b4i   & 68.85 & 73.91 & 44.55 & 88.58 & 76.98 & 90.13 & 97.16 & 95.94 & 70.98 & 77.95 \\
& rCLIP 6b2i   & \textbf{72.13} & 75.14 & \textbf{51.79} & 89.36 & 80.18 & 91.33 & \textbf{98.09} & \textbf{96.63} & 71.10 & \textbf{81.13} \\
\cmidrule(lr){2-12}
& r²CLIP 1b12i & 55.35 & 66.05 & 26.04 & 87.54 & 65.27 & 87.30 & 87.37 & 92.78 & 61.35 & 69.13 \\
& r²CLIP 3b4i  & 66.48 & 72.91 & 41.01 & 87.07 & 75.83 & 90.22 & 95.21 & 95.25 & 67.85 & 76.63 \\
& r²CLIP 6b2i  & 71.44 & \textbf{75.69} & 48.30 & \textbf{89.70} & \textbf{80.44} & \textbf{92.59} & 97.08 & 96.23 & \textbf{72.99} & 79.36 \\
\bottomrule
\end{tabular}%
}
\end{table*}

\noindent \textbf{\emph{Visual grounding.}} Table \ref{tab:grounding-results} reports grounding accuracy on RefCOCO, RefCOCO+, RefCOCOg, ReferItGame, and Flickr30k Entities following the HiVG-based adaptation. Unlike the other downstream tasks, grounding shows little difference between recursive and non-recursive models. Most recursive configurations remain within a few points of the distilled reference ViT-B and the OpenAI and DFN-2B checkpoints. On RefCOCO testB, rCLIP 6b2i reaches 78.73 compared with 78.85 for the distilled ViT-B, while on ReferItGame it achieves 69.41, slightly exceeding the three ViT-B references at 68.26--68.98. In contrast to classification and retrieval, the block/iteration count has little effect on grounding. For rCLIP on RefCOCO val, performance increases only from 80.08 at one block to 83.24 at six, while on several splits the one-block models remain within two to three points of the six-block variants. rCLIP and r²CLIP are also closely matched, with no consistent ordering between them. These results suggest that HiVG grounding is relatively insensitive to the degree of parameter sharing or to whether recurrence is applied to the vision tower alone or to both encoders, with even strongly recursive models remaining competitive with the non-recursive baselines.


\begin{table*}[h!]

  \centering
  \caption{Visual grounding accuracy at IoU $\geq$ 0.5 following the HiVG \cite{xiao2024hivg} adaptation protocol. Bold marks the best recursive model per column, underline the best overall. The remaining block/iteration configurations
  are reported in the Supplementary.}
  \label{tab:grounding-results}
  \begin{tabular}{p{4cm}|ccc|ccc|cc|c|c}
  \toprule
  \multirow{2}{*}{\makecell{CLIP Model}} & \multicolumn{3}{c|}{RefCOCO} &
  \multicolumn{3}{c|}{RefCOCO+} & \multicolumn{2}{c|}{RefCOCOg} & ReferIt & Flickr \\
  & val & testA & testB & val & testA & testB & val & test & test & test \\
  \midrule
  ViT-B/16 (OAI chkpt) & 83.99 & \underline{86.86} & 78.45 & \underline{73.96} & \underline{81.98} & 63.06 & 74.16 & \underline{74.39} & 68.62 & \underline{80.85} \\
  Teacher & 83.03 & 85.97 & 78.26 & 72.89 & 79.19 & 63.10 & 73.43 & 73.52 & 68.26 & 79.65 \\
  Ref. CLIP & \underline{84.31} & 86.68 & \underline{78.85} & 73.21 & 78.46 & \underline{64.74} & \underline{75.25} & 74.00 & 68.98 & 80.18 \\

 \cmidrule(lr){1-11}

  rCLIP 1b12i & 80.08 & 82.81 & 74.55 & 67.61 & 74.13 & 57.95 & 69.26 & 68.73 & 67.12 & 76.44 \\

  rCLIP 3b4i & 81.75 & 84.56 & 76.96 & 70.99 & 76.78 & 61.30 & 72.24 & \textbf{71.93} & 68.19 & 78.28 \\

  rCLIP 6b2i & \textbf{83.24} & \textbf{85.05} & \textbf{78.73} & 72.38 & 77.23 & 63.45 & 71.94 & 71.64 & \textbf{\underline{69.41}} & \textbf{79.40} \\

 \cmidrule(lr){1-11}

  r²CLIP 1b12i & 79.95 & 82.70 & 75.16 & 68.08 & 75.31 & 59.77 & 68.42 & 68.57 & 66.58 & 76.36 \\
  r²CLIP 3b4i & 82.04 & \textbf{85.05} & 76.22 & 71.06 & 77.32 & 62.57 & 72.10 & 71.27 & 68.24 & 78.53 \\
  r²CLIP 6b2i & 82.80 & 84.88 & 78.04 & \textbf{\underline{73.90}} & \textbf{78.95} & \textbf{64.10} & \textbf{72.94} & 71.78 & \textbf{\underline{69.41}} & 79.27 \\

  \bottomrule
  \end{tabular}
\end{table*}

\begin{table*}[h]
\centering
\caption{Zero-shot and post-DIET results. $D_f$ is forget-set and $D_t$ test-set
accuracy, averaged over five seeds with one randomly selected class removed per seed; lower
$D_f$ indicates more effective forgetting and higher $D_t$ better retention. In the post-DIET block, \textbf{bold}
marks the best recursive model and \underline{underline} the best model overall on average $D_t$. The remaining configurations are reported in the Supplementary.}
\label{tab:diet_results}
\resizebox{\textwidth}{!}{
\begin{tabular}{l cc cc cc cc}
\toprule
\multirow{2}{*}{\textbf{CLIP Model}}
& \multicolumn{2}{c}{\textbf{Flowers102}}
& \multicolumn{2}{c}{\textbf{Pets}}
& \multicolumn{2}{c}{\textbf{Food101}}
& \multicolumn{2}{c}{\textbf{Average}} \\
\cmidrule(lr){2-3}
\cmidrule(lr){4-5}
\cmidrule(lr){6-7}
\cmidrule(lr){8-9}
& $D_f$ (\%) & $D_t$ (\%)
& $D_f$ (\%) & $D_t$ (\%)
& $D_f$ (\%) & $D_t$ (\%)
& $D_f$ (\%) & $D_t$ (\%) \\
\midrule
\multicolumn{9}{l}{\textbf{Zero-shot}} \\
ViT-B/16 (OAI chkpt)
& $92.61{\pm}8.30$ & $70.56{\pm}0.12$
& $83.25{\pm}21.55$ & $89.12{\pm}0.37$
& $87.68{\pm}4.34$ & $85.85{\pm}0.05$
& $87.85$ & $81.84$ \\
Teacher
& $83.50{\pm}27.84$ & $80.35{\pm}0.16$
& $92.00{\pm}9.30$ & $93.34{\pm}0.23$
& $90.24{\pm}3.13$ & $87.77{\pm}0.05$
& $88.58$ & $87.15$ \\
Ref. CLIP
& $88.39{\pm}19.50$ & $66.72{\pm}0.13$
& $89.25{\pm}13.91$ & $92.18{\pm}0.29$
& $88.44{\pm}2.75$ & $80.23{\pm}0.04$
& $88.69$ & $79.71$ \\
 \cmidrule(lr){1-9}
rCLIP 1b12i
& $66.96{\pm}14.31$ & $49.22{\pm}0.25$
& $85.49{\pm}11.87$ & $87.20{\pm}0.22$
& $66.56{\pm}11.19$ & $58.37{\pm}0.10$
& $73.00$ & $64.93$ \\
rCLIP 3b4i
& $75.00{\pm}28.97$ & $60.70{\pm}0.13$
& $83.00{\pm}13.27$ & $88.66{\pm}0.26$
& $79.04{\pm}8.04$ & $71.05{\pm}0.07$
& $79.01$ & $73.47$ \\
rCLIP 6b2i
& $81.51{\pm}26.65$ & $63.44{\pm}0.17$
& $84.00{\pm}15.85$ & $90.65{\pm}0.25$
& $82.84{\pm}6.36$ & $74.78{\pm}0.05$
& $82.78$ & $76.29$ \\
 \cmidrule(lr){1-9}
r²CLIP 1b12i
& $68.85{\pm}21.92$ & $36.40{\pm}0.31$
& $78.98{\pm}18.49$ & $83.42{\pm}0.42$
& $71.24{\pm}4.53$ & $54.15{\pm}0.04$
& $73.02$ & $57.99$ \\
r²CLIP 3b4i
& $88.70{\pm}7.79$ & $57.14{\pm}0.13$
& $85.00{\pm}15.08$ & $89.26{\pm}0.19$
& $76.24{\pm}6.20$ & $68.43{\pm}0.06$
& $83.31$ & $71.61$ \\
r²CLIP 6b2i
& $81.88{\pm}24.64$ & $59.06{\pm}0.18$
& $84.50{\pm}14.96$ & $89.48{\pm}0.26$
& $79.76{\pm}8.21$ & $73.05{\pm}0.08$
& $82.05$ & $73.86$ \\
\midrule\midrule
\multicolumn{9}{l}{\textbf{After DIET}} \\
ViT-B/16 (OAI chkpt)
& $4.36{\pm}5.06$ & $56.60{\pm}2.06$
& $2.00{\pm}2.59$ & $73.79{\pm}9.93$
& $12.72{\pm}14.34$ & $72.80{\pm}4.00$
& $6.36$ & $67.73$ \\
Teacher
& $3.87{\pm}6.50$ & $74.24{\pm}3.46$
& $5.01{\pm}4.68$ & $77.18{\pm}4.75$
& $13.08{\pm}11.63$ & $59.15{\pm}11.14$
& $7.32$ & $\underline{70.19}$ \\
Ref. CLIP
& $2.60{\pm}3.16$ & $56.45{\pm}4.57$
& $3.00{\pm}5.35$ & $69.10{\pm}14.18$
& $14.12{\pm}12.58$ & $47.10{\pm}13.61$
& $6.57$ & $57.55$ \\
 \cmidrule(lr){1-9}
rCLIP 1b12i
& $4.29{\pm}6.27$ & $33.48{\pm}10.18$
& $12.58{\pm}12.10$ & $45.62{\pm}9.19$
& $9.32{\pm}9.26$ & $37.46{\pm}8.65$
& $8.73$ & $38.85$ \\
rCLIP 3b4i
& $7.20{\pm}8.16$ & $51.92{\pm}2.83$
& $22.72{\pm}36.72$ & $59.82{\pm}10.85$
& $15.52{\pm}6.62$ & $44.93{\pm}4.72$
& $15.15$ & $52.22$ \\
rCLIP 6b2i
& $1.84{\pm}2.06$ & $44.67{\pm}8.61$
& $4.02{\pm}2.44$ & $67.68{\pm}6.17$
& $9.56{\pm}5.12$ & $51.19{\pm}8.84$
& $5.14$ & $54.51$ \\
 \cmidrule(lr){1-9}
r²CLIP 1b12i
& $5.67{\pm}5.45$ & $27.73{\pm}4.16$
& $13.14{\pm}24.61$ & $56.63{\pm}13.02$
& $5.96{\pm}7.42$ & $31.77{\pm}3.09$
& $8.26$ & $38.71$ \\
r²CLIP 3b4i
& $5.15{\pm}8.67$ & $40.12{\pm}11.39$
& $13.03{\pm}12.00$ & $64.77{\pm}9.37$
& $9.40{\pm}7.32$ & $50.39{\pm}4.73$
& $9.19$ & $51.76$ \\
r²CLIP 6b2i
& $0.82{\pm}1.13$ & $44.30{\pm}11.42$
& $6.27{\pm}6.19$ & $68.28{\pm}11.80$
& $9.56{\pm}5.53$ & $52.27{\pm}7.47$
& $5.55$ & $\mathbf{54.95}$ \\
\bottomrule
\end{tabular}
}
\end{table*}

\noindent \textbf{\emph{Unlearning.}} Table \ref{tab:diet_results} reports zero-shot and post-unlearning forget-set ($D_f$) and test-set ($D_t$) accuracy on Flowers102, Pets, and Food101, averaged over five seeds with one randomly selected class removed per seed. Before unlearning, the distilled ViT-B is competitive with the OpenAI checkpoint, while the recursive models follow the expected ordering with the number of unique blocks. DIET is effective on every backbone: average forget-set accuracy falls to $5.1$--$15.2\%$ across the recursive configurations, comparable to $6.4$--$7.3\%$ for the non-recursive references. The models therefore differ mainly in how much test accuracy they retain. Averaged over the three datasets, $D_t$ falls by $22.2$ points for the distilled ViT-B versus $14.1$ for the OpenAI checkpoint, with the largest drop on Food101 ($80.2$ to $47.1$ versus $85.9$ to $72.8$). Since this behavior is already present in the non-recursive student, it does not appear to be caused by recursion. The DFN-2B teacher also loses more than the OpenAI checkpoint ($17.0$ points on average), suggesting that the pretraining corpus and shared learning-rate protocol may contribute alongside distillation and the smaller training set. Within the recursive families, models with more unique blocks generally retain accuracy better, although the trend varies across datasets. In Pets for example,  rCLIP 1b12i drops from $87.2$ to $45.6$, compared with $90.7$ to $67.7$ for rCLIP 6b2i and $92.2$ to $69.1$ for the non-recursive reference. Differences between rCLIP and r²CLIP are minor at matched configurations, with average post-DIET $D_t$ within $0.6$ points at every block count shown, indicating that text-tower recursion has little effect on unlearning behavior. 
\section{Conclusions}
\label{sec:conclusions}

In this work, we systematically studied recursive CLIP architectures based on depth-wise block reuse, considering both vision-only recurrence in rCLIP and joint vision--text recurrence in r²CLIP. Across a range of block and iteration configurations, we show that substantial recurrence can be introduced while preserving much of the transferable capability of CLIP. Increasing block reuse naturally reduces the number of distinct parameters, yielding a controllable trade-off between parameterization and representation quality. Our results further show that the vision and text encoders respond differently to extensive weight sharing, with the cost of text-tower recurrence becoming small as more distinct blocks are retained.

Our distillation experiments show that strong CLIP representations can be transferred to recursive students despite substantial parameter sharing. Beyond zero-shot classification and cross-modal retrieval, evaluation on few-shot adaptation, visual grounding, and concept unlearning reveals that the effect of recurrence depends on the downstream task. Recursive models remain highly competitive for grounding and benefit strongly from few-shot adaptation, while heavily shared configurations can be more sensitive during unlearning. Block reuse also reduces the number of distinct LoRA adapters required for downstream adaptation. Overall, our results establish recurrent block reuse as a practical and flexible design space for CLIP, with parameter efficiency arising naturally from the shared architecture.

Our work has some limitations. Most experiments use a single training run, so small performance differences should be interpreted cautiously. We also don't evaluate alternative methods for tuning recursive models, such as block-specific iteration counts~\cite{jacobs2026block}, which remain a direction for future work. Finally, the downstream adaptation protocols are adopted from methods originally designed for non-recursive models. Their performance may therefore be sensitive to hyperparameter choices, and further tuning specifically for recursive models could lead to improvements.

\section*{Acknowledgements}

We gratefully acknowledge Polish high-performance computing infrastructure PLGrid (HPC Center: ACK Cyfronet AGH) for providing computer facilities and support within computational grant no. PLG/2025/018809 

{
    \small
    \bibliographystyle{ieeenat_fullname}
    \bibliography{main}

@String(CVPR  = {Proceedings of the IEEE/CVF Conference on Computer Vision and Pattern Recognition})

@String(CVPRW = {Proceedings of the IEEE/CVF Conference on Computer Vision and Pattern Recognition Workshops})

@String(ICCV  = {Proceedings of the IEEE/CVF International Conference on Computer Vision})

@String(ECCV  = {Proceedings of the European Conference on Computer Vision})

@String(WACV  = {Proceedings of the IEEE/CVF Winter Conference on Applications of Computer Vision})

@String(NIPS  = {Advances in Neural Information Processing Systems})

@String(ICLR  = {International Conference on Learning Representations})

@String(ICML  = {International Conference on Machine Learning})

@String(AAAI  = {Proceedings of the AAAI Conference on Artificial Intelligence})

@String(ACMMM = {Proceedings of the ACM International Conference on Multimedia})

@String(ACL   = {Proceedings of the Annual Meeting of the Association for Computational Linguistics})

@String(EMNLP = {Proceedings of the Conference on Empirical Methods in Natural Language Processing})

@String(IJCV  = {International Journal of Computer Vision})

@String(TACL  = {Transactions of the Association for Computational Linguistics})

@inproceedings{radford2021learning,
  title     = {Learning Transferable Visual Models from Natural Language Supervision},
  author    = {Radford, Alec and Kim, Jong Wook and Hallacy, Chris and Ramesh, Aditya and Goh, Gabriel and Agarwal, Sandhini and Sastry, Girish and Askell, Amanda and Mishkin, Pamela and Clark, Jack and Krueger, Gretchen and Sutskever, Ilya},
  booktitle = ICML,
  pages     = {8748--8763},
  year      = {2021}
}

@inproceedings{jia2021scaling,
  title     = {Scaling Up Visual and Vision-Language Representation Learning with Noisy Text Supervision},
  author    = {Jia, Chao and Yang, Yinfei and Xia, Ye and Chen, Yi-Ting and Parekh, Zarana and Pham, Hieu and Le, Quoc and Sung, Yun-Hsuan and Li, Zhen and Duerig, Tom},
  booktitle = ICML,
  pages     = {4904--4916},
  year      = {2021}
}

@inproceedings{zhai2023sigmoid,
  title     = {Sigmoid Loss for Language Image Pre-Training},
  author    = {Zhai, Xiaohua and Mustafa, Basil and Kolesnikov, Alexander and Beyer, Lucas},
  booktitle = ICCV,
  pages     = {11975--11986},
  year      = {2023}
}

@inproceedings{fang2024data,
  title     = {Data Filtering Networks},
  author    = {Fang, Alex and Jose, Albin Madappally and Jain, Amit and Schmidt, Ludwig and Toshev, Alexander and Shankar, Vaishaal},
  booktitle = ICLR,
  year      = {2024}
}

@inproceedings{yang2024clip,
  title     = {{CLIP-KD}: An Empirical Study of {CLIP} Model Distillation},
  author    = {Yang, Chuanguang and An, Zhulin and Huang, Libo and Bi, Junyu and Yu, Xinqiang and Yang, Han and Diao, Boyu and Xu, Yongjun},
  booktitle = CVPR,
  pages     = {15952--15962},
  year      = {2024}
}

@inproceedings{wu2023tinyclip,
  title     = {{TinyCLIP}: {CLIP} Distillation via Affinity Mimicking and Weight Inheritance},
  author    = {Wu, Kan and Peng, Houwen and Zhou, Zhenghong and Xiao, Bin and Liu, Mengchen and Yuan, Lu and Xuan, Hong and Valenzuela, Michael and Chen, Xi Stephen and Wang, Xinggang and Chao, Hongyang and Hu, Han},
  booktitle = ICCV,
  pages     = {21970--21980},
  year      = {2023}
}

@inproceedings{vasu2024mobileclip,
  title     = {{MobileCLIP}: Fast Image-Text Models through Multi-Modal Reinforced Training},
  author    = {Vasu, Pavan Kumar Anasosalu and Pouransari, Hadi and Faghri, Fartash and Vemulapalli, Raviteja and Tuzel, Oncel},
  booktitle = CVPR,
  pages     = {15963--15974},
  year      = {2024}
}

@inproceedings{zanella2024low,
  title     = {Low-Rank Few-Shot Adaptation of Vision-Language Models},
  author    = {Zanella, Maxime and Ben Ayed, Ismail},
  booktitle = CVPRW,
  pages     = {1593--1603},
  year      = {2024}
}

@inproceedings{wu2025patch,
  title={Patch Ranking: Token Pruning as Ranking Prediction for Efficient CLIP},
  author={Wu, Cheng-En and Lin, Jinhong and Hu, Yu Hen and Morgado, Pedro},
  booktitle={2025 IEEE/CVF Winter Conference on Applications of Computer Vision (WACV)},
  pages={5842--5851},
  year={2025},
  organization={IEEE}
}

@inproceedings{fahim2026rmma,
  title     = {{R-MMA}: Enhancing Vision-Language Models with Recurrent Adapters for Few-Shot and Cross-Domain Generalization},
  author    = {Fahim, Md and Ishmam, Md Farhan and Hossain, Mir Sazzat and Amin, M Ashraful and Ali, Amin Ahsan and Rahman, AKM Mahbubur},
  booktitle = WACV,
  pages     = {6496--6506},
  year      = {2026}
}

@inproceedings{kunananthaseelan2026diet,
  title     = {{DIET}: Machine Unlearning on a Data-Diet},
  author    = {Kunananthaseelan, Nilakshan and Wu, Jing and Le, Trung and Haffari, Gholamreza and Harandi, Mehrtash},
  booktitle = AAAI,
  volume    = {40},
  number    = {27},
  pages     = {22698--22706},
  year      = {2026}
}

@inproceedings{xiao2024hivg,
  title     = {{HiVG}: Hierarchical Multimodal Fine-Grained Modulation for Visual Grounding},
  author    = {Xiao, Linhui and Yang, Xiaoshan and Peng, Fang and Wang, Yaowei and Xu, Changsheng},
  booktitle = ACMMM,
  pages     = {5460--5469},
  year      = {2024}
}

@inproceedings{gao2023learning,
  title     = {Learning to Jointly Share and Prune Weights for Grounding Based Vision and Language Models},
  author    = {Gao, Shangqian and Uzkent, Burak and Shen, Yilin and Huang, Heng and Jin, Hongxia},
  booktitle = ICLR,
  year      = {2023}
}

@inproceedings{yu2016modeling,
  title     = {Modeling Context in Referring Expressions},
  author    = {Yu, Licheng and Poirson, Patrick and Yang, Shan and Berg, Alexander C. and Berg, Tamara L.},
  booktitle = ECCV,
  pages     = {69--85},
  year      = {2016}
}

@inproceedings{mao2016generation,
  title     = {Generation and Comprehension of Unambiguous Object Descriptions},
  author    = {Mao, Junhua and Huang, Jonathan and Toshev, Alexander and Camburu, Oana and Yuille, Alan L. and Murphy, Kevin},
  booktitle = CVPR,
  pages     = {11--20},
  year      = {2016}
}

@inproceedings{kazemzadeh2014referitgame,
  title     = {{ReferItGame}: Referring to Objects in Photographs of Natural Scenes},
  author    = {Kazemzadeh, Sahar and Ordonez, Vicente and Matten, Mark and Berg, Tamara},
  booktitle = EMNLP,
  pages     = {787--798},
  year      = {2014}
}

@inproceedings{plummer2015flickr30k,
  title     = {{Flickr30k} Entities: Collecting Region-to-Phrase Correspondences for Richer Image-to-Sentence Models},
  author    = {Plummer, Bryan A. and Wang, Liwei and Cervantes, Chris M. and Caicedo, Juan C. and Hockenmaier, Julia and Lazebnik, Svetlana},
  booktitle = ICCV,
  pages     = {2641--2649},
  year      = {2015}
}

@inproceedings{vaswani2017attention,
  title     = {Attention Is All You Need},
  author    = {Vaswani, Ashish and Shazeer, Noam and Parmar, Niki and Uszkoreit, Jakob and Jones, Llion and Gomez, Aidan N. and Kaiser, {\L}ukasz and Polosukhin, Illia},
  booktitle = NIPS,
  volume    = {30},
  pages     = {5998--6008},
  year      = {2017}
}

@inproceedings{dehghani2018universal,
  title     = {Universal Transformers},
  author    = {Dehghani, Mostafa and Gouws, Stephan and Vinyals, Oriol and Uszkoreit, Jakob and Kaiser, Lukasz},
  booktitle = ICLR,
  year      = {2019}
}

@inproceedings{Lan2020ALBERT,
  title     = {{ALBERT}: A Lite {BERT} for Self-Supervised Learning of Language Representations},
  author    = {Lan, Zhenzhong and Chen, Mingda and Goodman, Sebastian and Gimpel, Kevin and Sharma, Piyush and Soricut, Radu},
  booktitle = ICLR,
  year      = {2020}
}

@inproceedings{shen2022sliced,
  title     = {Sliced Recursive Transformer},
  author    = {Shen, Zhiqiang and Liu, Zechun and Xing, Eric},
  booktitle = ECCV,
  pages     = {727--744},
  year      = {2022}
}

@inproceedings{zhang2022minivit,
  title     = {{MiniViT}: Compressing Vision Transformers with Weight Multiplexing},
  author    = {Zhang, Jinnian and Peng, Houwen and Wu, Kan and Liu, Mengchen and Xiao, Bin and Fu, Jianlong and Yuan, Lu},
  booktitle = CVPR,
  pages     = {12145--12154},
  year      = {2022}
}

@inproceedings{giannou2023looped,
  title     = {Looped Transformers as Programmable Computers},
  author    = {Giannou, Angeliki and Rajput, Shashank and Sohn, Jy-yong and Lee, Kangwook and Lee, Jason D. and Papailiopoulos, Dimitris},
  booktitle = ICML,
  pages     = {11398--11442},
  year      = {2023}
}

@inproceedings{yang2023looped,
  title     = {Looped Transformers Are Better at Learning Learning Algorithms},
  author    = {Yang, Liu and Lee, Kangwook and Nowak, Robert and Papailiopoulos, Dimitris},
  booktitle = ICLR,
  year      = {2024}
}

@inproceedings{saunshi2025reasoning,
  title     = {Reasoning with Latent Thoughts: On the Power of Looped Transformers},
  author    = {Saunshi, Nikunj and Dikkala, Nishanth and Li, Zhiyuan and Kumar, Sanjiv and Reddi, Sashank J.},
  booktitle = ICLR,
  year      = {2025}
}

@inproceedings{bae2025relaxed,
  title     = {Relaxed Recursive Transformers: Effective Parameter Sharing with Layer-Wise {LoRA}},
  author    = {Bae, Sangmin and Fisch, Adam and Harutyunyan, Hrayr and Ji, Ziwei and Kim, Seungyeon and Schuster, Tal},
  booktitle = ICLR,
  year      = {2025}
}

@inproceedings{geiping2025scaling,
  title     = {Scaling Up Test-Time Compute with Latent Reasoning: A Recurrent Depth Approach},
  author    = {Geiping, Jonas and McLeish, Sean and Jain, Neel and Kirchenbauer, John and Singh, Siddharth and Bartoldson, Brian R. and Kailkhura, Bhavya and Bhatele, Abhinav and Goldstein, Tom},
  booktitle = NIPS,
  volume    = {38},
  pages     = {41340--41391},
  year      = {2025}
}

@inproceedings{jacobs2026block,
  title     = {Block-Recurrent Dynamics in Vision Transformers},
  author    = {Jacobs, Mozes and Fel, Thomas and Hakim, Richard and Brondetta, Alessandra and Ba, Demba and Keller, T. Anderson},
  booktitle = ICLR,
  year      = {2026}
}

@misc{byra2026bvit,
  title         = {{bViT}: Investigating Single-Block Recurrence in Vision Transformers for Image Recognition},
  author        = {Byra, Michal and Olszowiec, Pawel and Stefanski, Grzegorz and Gruszczynski, Grzegorz and Presta, Alberto},
  year          = {2026},
  eprint        = {2605.10661},
  archivePrefix = {arXiv},
  primaryClass  = {cs.CV}
}

@article{hinton2015distilling,
  title   = {Distilling the Knowledge in a Neural Network},
  author  = {Hinton, Geoffrey and Vinyals, Oriol and Dean, Jeff},
  journal = {arXiv preprint arXiv:1503.02531},
  year    = {2015}
}

@inproceedings{romero2015fitnets,
  title     = {{FitNets}: Hints for Thin Deep Nets},
  author    = {Romero, Adriana and Ballas, Nicolas and Kahou, Samira Ebrahimi and Chassang, Antoine and Gatta, Carlo and Bengio, Yoshua},
  booktitle = ICLR,
  year      = {2015}
}

@inproceedings{loshchilov2018decoupled,
  title     = {Decoupled Weight Decay Regularization},
  author    = {Loshchilov, Ilya and Hutter, Frank},
  booktitle = ICLR,
  year      = {2019}
}

@inproceedings{sharma2018conceptual,
  title     = {Conceptual Captions: A Cleaned, Hypernymed, Image Alt-Text Dataset for Automatic Image Captioning},
  author    = {Sharma, Piyush and Ding, Nan and Goodman, Sebastian and Soricut, Radu},
  booktitle = ACL,
  pages     = {2556--2565},
  year      = {2018}
}

@inproceedings{changpinyo2021conceptual,
  title     = {Conceptual {12M}: Pushing Web-Scale Image-Text Pre-Training to Recognize Long-Tail Visual Concepts},
  author    = {Changpinyo, Soravit and Sharma, Piyush and Ding, Nan and Soricut, Radu},
  booktitle = CVPR,
  pages     = {3558--3568},
  year      = {2021}
}

@inproceedings{lin2014mscoco,
  title     = {Microsoft {COCO}: Common Objects in Context},
  author    = {Lin, Tsung-Yi and Maire, Michael and Belongie, Serge and Hays, James and Perona, Pietro and Ramanan, Deva and Doll{\'a}r, Piotr and Zitnick, C. Lawrence},
  booktitle = ECCV,
  pages     = {740--755},
  year      = {2014}
}

@article{young2014image,
  title   = {From Image Descriptions to Visual Denotations: New Similarity Metrics for Semantic Inference over Event Descriptions},
  author  = {Young, Peter and Lai, Alice and Hodosh, Micah and Hockenmaier, Julia},
  journal = TACL,
  volume  = {2},
  pages   = {67--78},
  year    = {2014}
}

@inproceedings{karpathy2015deep,
  title     = {Deep Visual-Semantic Alignments for Generating Image Descriptions},
  author    = {Karpathy, Andrej and Fei-Fei, Li},
  booktitle = CVPR,
  pages     = {3128--3137},
  year      = {2015}
}

@article{russakovsky2015imagenet,
  title   = {{ImageNet} Large Scale Visual Recognition Challenge},
  author  = {Russakovsky, Olga and Deng, Jia and Su, Hao and Krause, Jonathan and Satheesh, Sanjeev and Ma, Sean and Huang, Zhiheng and Karpathy, Andrej and Khosla, Aditya and Bernstein, Michael and Berg, Alexander C. and Fei-Fei, Li},
  journal = IJCV,
  volume  = {115},
  number  = {3},
  pages   = {211--252},
  year    = {2015}
}

@inproceedings{recht2019imagenet,
  title     = {Do {ImageNet} Classifiers Generalize to {ImageNet}?},
  author    = {Recht, Benjamin and Roelofs, Rebecca and Schmidt, Ludwig and Shankar, Vaishaal},
  booktitle = ICML,
  pages     = {5389--5400},
  year      = {2019}
}

@inproceedings{hendrycks2021many,
  title     = {The Many Faces of Robustness: A Critical Analysis of Out-of-Distribution Generalization},
  author    = {Hendrycks, Dan and Basart, Steven and Mu, Norman and Kadavath, Saurav and Wang, Frank and Dorundo, Evan and Desai, Rahul and Zhu, Tyler and Parajuli, Samyak and Guo, Mike and Song, Dawn and Steinhardt, Jacob and Gilmer, Justin},
  booktitle = ICCV,
  pages     = {8340--8349},
  year      = {2021}
}

@inproceedings{wang2019learning,
  title     = {Learning Robust Global Representations by Penalizing Local Predictive Power},
  author    = {Wang, Haohan and Ge, Songwei and Lipton, Zachary and Xing, Eric P.},
  booktitle = NIPS,
  volume    = {32},
  pages     = {10506--10518},
  year      = {2019}
}

@article{zhou2022coop,
    title={Learning to Prompt for Vision-Language Models},
    author={Zhou, Kaiyang and Yang, Jingkang and Loy, Chen Change and Liu, Ziwei},
    journal={International Journal of Computer Vision (IJCV)},
    year={2022}
}
}

\clearpage
\appendix
\maketitlesupplementary

\section{Distillation loss details}
\label{app:loss-defs}

This section gives the formal definitions of the distillation objectives
summarized in Sect.~\ref{method:3.2}. Throughout, $\mathcal{B} = \{(I_k, T_k)\}_{k=1}^{|\mathcal{B}|}$
is a mini-batch of image--text pairs, superscripts $\mathbf{T}$ and $\mathbf{S}$
denote teacher and student, and $v_k$, $s_k$ denote $\ell_2$-normalized image and
text embeddings.

\subsection{CLIP Contrastive Loss}

Given a visual encoder $f_i$ and text encoder $f_t$ produce $v_k = f_i(I_k)$ and
$s_k = f_t(T_k)$, CLIP uses an InfoNCE-based objective that pulls paired
embeddings together and pushes unpaired ones apart. With the image as anchor,
\begin{equation}
\mathcal{L}_{I \to T} = -\log \frac{\exp(v_k \cdot s_k / \tau)}
{\sum_{b=1}^{|\mathcal{B}|} \exp(v_k \cdot s_b / \tau)},
\end{equation}
and symmetrically, with the text as anchor,
\begin{equation}
\mathcal{L}_{T \to I} = -\log \frac{\exp(s_k \cdot v_k / \tau)}
{\sum_{b=1}^{|\mathcal{B}|} \exp(s_k \cdot v_b / \tau)},
\end{equation}
where $\cdot$ is the dot product and $\tau$ a learnable temperature. The full
CLIP loss is the average of both directions:
\begin{equation}
\mathcal{L}_{\text{CLIP}} = \tfrac{1}{2}\left(\mathcal{L}_{I \to T}
+ \mathcal{L}_{T \to I}\right).
\end{equation}

\subsection{Feature Distillation (FD)}

FD directly aligns student embeddings to the teacher's by minimizing the mean
squared error between their $\ell_2$-normalized visual and text features:
\begin{equation}
\mathcal{L}_{\text{FD}} = \frac{1}{|\mathcal{B}|} \sum_{k=1}^{|\mathcal{B}|}
\left( \left\| v_k^{\mathbf{T}} - v_k^{\mathbf{S}} \right\|_2^2
+ \left\| s_k^{\mathbf{T}} - s_k^{\mathbf{S}} \right\|_2^2 \right).
\end{equation}
When embedding dimensions differ, a linear projection head maps student
features to the teacher dimension; all students considered here preserve the
ViT-B/16 embedding width, so no such head is instantiated. Intuitively,
forcing the student's features toward the teacher's shrinks the performance
gap directly.

\subsection{Contrastive Relational Distillation (CRD)}

CRD transfers the teacher's structured contrastive distribution rather than raw
features. For image $k$ as anchor, the teacher and student image-to-text
distributions are
\begin{equation}
\begin{aligned}
p_k^{\mathbf{T}}[j] &= \frac{\exp(v_k^{\mathbf{T}} \cdot s_j^{\mathbf{T}}
/ \tau^{\mathbf{T}})}{\sum_{b=1}^{|\mathcal{B}|} \exp(v_k^{\mathbf{T}}
\cdot s_b^{\mathbf{T}} / \tau^{\mathbf{T}})}, \\[6pt]
p_k^{\mathbf{S}}[j] &= \frac{\exp(v_k^{\mathbf{S}} \cdot s_j^{\mathbf{S}}
/ \tau^{\mathbf{S}})}{\sum_{b=1}^{|\mathcal{B}|} \exp(v_k^{\mathbf{S}}
\cdot s_b^{\mathbf{S}} / \tau^{\mathbf{S}})},
\end{aligned}
\end{equation}
with analogous text-to-image distributions $q_k^{\mathbf{T}}$ and
$q_k^{\mathbf{S}}$. The student mimics the teacher via KL divergence in both
directions:
\begin{equation}
\mathcal{L}_{\text{CRD}} = \frac{1}{|\mathcal{B}|} \sum_{k=1}^{|\mathcal{B}|}
\sum_{j=1}^{|\mathcal{B}|} \left( p_k^{\mathbf{T}}[j]
\log \frac{p_k^{\mathbf{T}}[j]}{p_k^{\mathbf{S}}[j]}
+ q_k^{\mathbf{T}}[j] \log \frac{q_k^{\mathbf{T}}[j]}{q_k^{\mathbf{S}}[j]}
\right).
\end{equation}
This encourages the student to reproduce the teacher's relational structure over
the batch.

\subsection{Interactive Contrastive Learning (ICL)}

ICL couples the two encoders by treating the student embedding as an anchor
contrasted against teacher embeddings. Using the student image embedding
$v_k^{\mathbf{S}}$ against teacher text embeddings
$\{s_b^{\mathbf{T}}\}_{b=1}^{|\mathcal{B}|}$,
\begin{equation}
\mathcal{L}_{\text{ICL}, I \to T} = -\log \frac{\exp(v_k^{\mathbf{S}}
\cdot s_k^{\mathbf{T}} / \tau)}{\sum_{b=1}^{|\mathcal{B}|}
\exp(v_k^{\mathbf{S}} \cdot s_b^{\mathbf{T}} / \tau)},
\end{equation}
and symmetrically, using the student text embedding against teacher image
embeddings,
\begin{equation}
\mathcal{L}_{\text{ICL}, T \to I} = -\log \frac{\exp(s_k^{\mathbf{S}}
\cdot v_k^{\mathbf{T}} / \tau)}{\sum_{b=1}^{|\mathcal{B}|}
\exp(s_k^{\mathbf{S}} \cdot v_b^{\mathbf{T}} / \tau)}.
\end{equation}
The total ICL loss averages both directions:
\begin{equation}
\mathcal{L}_{\text{ICL}} = \tfrac{1}{2}\left(\mathcal{L}_{\text{ICL}, I \to T}
+ \mathcal{L}_{\text{ICL}, T \to I}\right).
\end{equation}
Minimizing $\mathcal{L}_{\text{ICL}}$ maximizes a lower bound on the mutual
information between teacher and student representations, letting the student
learn implicitly from the teacher's cross-modal embedding space.

\subsection{Activation Matching (AM)}
\label{app:am}

We give here the full definition of the objective introduced in Sect.~\ref{method:3.2},
whose behavior under recursion is analyzed in Sect.~\ref{sec:loss-choice}. Unlike the
objectives above, which act on the final embeddings, AM~\cite{romero2015fitnets}
supervises intermediate representations along the depth of the vision tower,
and is applied to the vision tower only.

\paragraph{Teacher trajectory.}
Let $\mathcal{T}$ denote the frozen teacher vision tower. For an image $I_k$ it
embeds patches into a token sequence of $P+1$ tokens ($P = 196$ patch tokens
and one \texttt{[CLS]} token) of width $d_T$, and applies its $L = 12$ residual
blocks,
\begin{equation}
  H^{\ell} = \mathcal{B}_T^{\ell}\!\left(H^{\ell-1}\right)
           = H^{\ell-1} + \Delta_T^{\ell}\!\left(H^{\ell-1}\right),
  \qquad \ell = 1,\dots,L,
\end{equation}
with $H^{0}$ the teacher patch embedding. We refer to
$(H^{\ell})_{\ell=1}^{L}$ as the teacher's residual-stream trajectory; these
are raw block outputs, taken before the final layer normalization and before
the CLIP projection head.

\paragraph{Student trajectory.}
The corresponding student states are obtained by unrolling the recursion of
Eq.~\eqref{eq:recursion}. Writing $A^{m,i} = B_i(A^{m,i-1})$ for the state after the $i$-th
distinct block within iteration $m$, with $A^{m,0} = A^{m-1}$ and
$A^{m,N} = A^{m}$, the global execution index of Eq.~(\ref{eq:unroll}),
$\ell = (m-1)N + i$, lets us write $A^{\ell} \equiv A^{m,i}$. Because all
configurations satisfy $N \cdot M = L$, student and teacher trajectories have
equal length and the alignment is index-preserving. As noted in the main text,
block $B_i$ is executed at $\ell \in \{i,\, N+i,\, \dots,\, (M-1)N+i\}$, so
one parameter set is supervised against $M$ teacher states spaced $N$ layers
apart; the non-recursive reference is the special case $N = L$, $M = 1$, in
which each parameter set receives exactly one target.

\paragraph{Objective.}
For a mini-batch $\mathcal{B} = \{I_k\}_{k=1}^{|\mathcal{B}|}$,
\begin{equation}
  \mathcal{L}_{\mathrm{AM}}
  = \frac{1}{L\,|\mathcal{B}|\,(P{+}1)\,d_T}
    \sum_{\ell=1}^{L} \sum_{k=1}^{|\mathcal{B}|}
    \left\| W A^{\ell}_k - \mathrm{sg}\!\left[H^{\ell}_k\right] \right\|_F^{2},
  \label{eq:am}
\end{equation}
where $\|\cdot\|_F$ is the Frobenius norm and $\mathrm{sg}[\cdot]$ the
stop-gradient (the teacher is frozen and evaluated without gradient tracking).
$W \in \mathbb{R}^{d_T \times d_S}$ is a learnable bias-free linear map
instantiated only when student and teacher widths differ; all students
considered here preserve the ViT-B/16 width, so $W = I$ throughout and no
additional parameters are introduced. The comparison is made on unnormalized
hidden states over the full token sequence, with \texttt{[CLS]} and patch
tokens weighted equally.

\paragraph{Scale.}
Averaging over $\ell$ and over all elements makes
Eq.~(\ref{eq:am}) invariant to trajectory length and hidden width, but places
it on a substantially smaller numerical scale than the output-level losses.
The weights examined for AM therefore span a wider range than those for the
other components; see Sect.~\ref{app:am-weight}.

\paragraph{Overall objective.}
The full training objective combines the base task with the selected
distillation terms,
\begin{equation}
  \mathcal{L} = \mathcal{L}_{\text{CLIP}}
  + \lambda_{\text{CRD}} \mathcal{L}_{\text{CRD}}
  + \lambda_{\text{ICL}} \mathcal{L}_{\text{ICL}}
  + \lambda_{\text{FD}} \mathcal{L}_{\text{FD}}
  + \alpha \mathcal{L}_{\mathrm{AM}},
\end{equation}
with $\alpha = 0$ in all main experiments following Sect.~\ref{sec:loss-choice}, and
$\lambda_{\text{CRD}} = 1$, $\lambda_{\text{ICL}} = 0$,
$\lambda_{\text{FD}} = 10^4$ following Sect.~\ref{app:loss-weighting}.

\section{Extended distillation objective results}
\label{app:loss}

This section supports Sect.~\ref{sec:loss-choice}. We report the component ablation on all four
zero-shot evaluation sets (Sect.~\ref{app:loss-composition}), the sensitivity
of Activation Matching to its weight (Sect.~\ref{app:am-weight}), and the
weighting sweep over the retained components (Sect.~\ref{app:loss-weighting}).
All runs train on CC3M only, for the same budget and with the same optimizer
settings as the main distillation runs, and use a single seed. Configurations
are compared by ImageNet accuracy; the three distribution-shifted variants are
reported for completeness. The recursive backbone throughout is
rCLIP~1b12i, the most constrained configuration in our grid and the one whose
performance we most seek to improve.

\subsection{Component ablation}
\label{app:loss-composition}

Table~\ref{tab:loss-composition-full} extends Table~\ref{tab:loss-composition} of the main text with
ImageNet-V2 and ImageNet-Sketch. The ordering discussed in Sect.~\ref{sec:loss-choice} is
preserved on every evaluation set: FD is the strongest individual component
on both backbones, CRD and ICL fall well short of the full composition, and AM
inverts between backbones, moving from second place on the reference to last
on the recursive student.

Two details are visible only in the extended table. First, the substitution of
the full composition by FD alone is close to free on the reference (37.3 to
37.0 on ImageNet, and in fact slightly favorable on ImageNet-Sketch, 27.0 to
27.9), but costs 2.1 points on the recursive student; the auxiliary components
contribute more under weight sharing. Second, AM on the reference backbone is
the single best run on ImageNet-R (47.5), exceeding even the full composition
(46.9), despite ranking only second on ImageNet. Intermediate supervision thus
appears to transfer robustness features that the output-level objectives do
not capture --- which makes its collapse on the recursive backbone, where it is
the weakest component on all four sets including ImageNet-R (17.0 against 33.2
for the full composition), the more striking.

\begin{table}[t]
\centering
\setlength{\tabcolsep}{4pt}
\small
\begin{tabular}{llcccc}
\toprule
Model & Components & IN & IN-V2 & IN-R & IN-S \\
\midrule
Ref. CLIP & CRD + ICL + FD & 37.3 & 32.5 & 46.9 & 27.0 \\
Ref. CLIP & FD             & 37.0 & 31.7 & 46.3 & 27.9 \\
Ref. CLIP & CRD            & 26.2 & 22.4 & 28.8 & 14.2 \\
Ref. CLIP & ICL            & 19.3 & 16.4 & 22.3 &  9.8 \\
Ref. CLIP & AM             & 31.0 & 26.5 & 47.5 & 23.9 \\
\midrule
rCLIP 1b12i & CRD + ICL + FD & 30.3 & 26.0 & 33.2 & 16.3 \\
rCLIP 1b12i & FD             & 28.2 & 24.3 & 29.2 & 14.4 \\
rCLIP 1b12i & CRD            & 22.4 & 18.8 & 22.6 &  9.3 \\
rCLIP 1b12i & ICL            & 18.4 & 15.2 & 19.6 &  7.2 \\
rCLIP 1b12i & AM             & 15.5 & 13.1 & 17.0 &  5.9 \\
\bottomrule
\end{tabular}
\caption{Distillation loss components evaluated in isolation, with the
reference composition of \cite{yang2024clip} as the top row of each block.
Extended version of Table~\ref{tab:loss-composition}. All runs additionally include the primary CLIP
contrastive loss at weight 1; CRD, ICL and FD use the weights of
\cite{yang2024clip} ($1$, $1$ and $2000$), and AM uses $\alpha = 2000$.
Values are top-1 accuracy (\%), single seed, CC3M, distilled from a DFN-2B
ViT-B/16 teacher.}
\label{tab:loss-composition-full}
\end{table}

\subsection{Sensitivity of Activation Matching to its weight}
\label{app:am-weight}

Because $\mathcal{L}_{\mathrm{AM}}$ sits on a smaller numerical scale than the
output-level objectives, a natural concern is that its poor showing on the
recursive backbone reflects an inadequate weight rather than a property of the
architecture. Table~\ref{tab:am-weight} addresses this by evaluating AM at
$\alpha = 1$ and $\alpha = 2000$ on both backbones.

The two backbones respond in opposite ways. On the reference, raising the
weight by three orders of magnitude improves ImageNet accuracy by 10.6 points
(20.4 to 31.0), with gains on all shifted variants: the objective is
weight-sensitive and clearly under-weighted at $\alpha = 1$. On the recursive
student the same change moves accuracy by 0.2 points in the opposite direction
(15.7 to 15.5), and by at most 1.1 points on any evaluation set. AM is
therefore not merely mis-weighted under recursion; across a $2000\times$ range
of weights it provides essentially no usable signal, while the identical sweep
on a non-recursive student of the same width and depth yields a large
improvement. This is what we would expect if the constraint imposed by
Eq.~(\ref{eq:am}) is unsatisfiable for a looped block rather than simply weak.

\begin{table}[t]
\centering
\setlength{\tabcolsep}{4pt}
\small
\begin{tabular}{llcccc}
\toprule
Model & $\alpha$ & IN & IN-V2 & IN-R & IN-S \\
\midrule
Ref. CLIP   & 1    & 20.4 & 17.6 & 23.0 & 10.3 \\
Ref. CLIP   & 2000 & 31.0 & 26.5 & 47.5 & 23.9 \\
\midrule
rCLIP 1b12i & 1    & 15.7 & 13.0 & 15.9 &  5.6 \\
rCLIP 1b12i & 2000 & 15.5 & 13.1 & 17.0 &  5.9 \\
\bottomrule
\end{tabular}
\caption{Activation Matching weight sensitivity. All runs use the primary CLIP
contrastive loss at weight 1 and AM as the only distillation term. Raising
$\alpha$  substantially improves the non-recursive baseline, with no visible effect in recursive one. Values are top-1 accuracy
(\%), single seed, CC3M, distilled from a DFN-2B ViT-B/16 teacher.}
\label{tab:am-weight}
\end{table}

\subsection{Loss weighting}
\label{app:loss-weighting}

Retaining CRD, ICL and FD, we next vary their weights on the same recursive
backbone. Results are given in Table~\ref{tab:loss-weights}.

Holding CRD and ICL at unit weight, accuracy increases monotonically with the
FD weight over the range examined --- $26.7$, $27.9$, $30.3$, $31.7$ and
$32.0$ for weights of $500$ to $5000$ --- with diminishing returns at the upper
end. We did not observe a turning point: performance was still improving at
$10^4$, where the search was stopped for budget reasons, so the optimum is
bounded from below but not bracketed.

The contribution of ICL is small and non-monotone once FD is heavily weighted. At an FD
weight of $5000$, ICL weights of $2$, $1$, $0.5$ and $0$ yield $31.3$, $32.0$,
$32.7$ and $32.5$ respectively, and at an FD weight of $10^4$ removing ICL
entirely is preferable to retaining it at half weight ($33.8$ against $33.6$).
By contrast, the CRD weight has little influence across the range examined,
varying accuracy by at most $0.6$ points between weights of $0.5$ and $2$. We
therefore adopt CRD $=1$, ICL $=0$ and FD $=10^4$, which attains $33.8$
ImageNet top-1 and is likewise the strongest configuration on ImageNet-V2 and
ImageNet-Sketch; ImageNet-R favors a different point (CRD $=2$, ICL $=0.5$,
FD $=5000$, at $36.8$), which we did not select as our selection criterion is
ImageNet top-1.

The selected configuration was verified on the non-recursive backbone before
being adopted for the main experiments. It raises Reference CLIP from $37.3$ to
$47.7$ ImageNet top-1 under the same budget, a larger absolute gain than on the
recursive student. The recipe used throughout the paper is therefore not tuned
in favor of the recursive models: all students, recursive and non-recursive,
are distilled under the objective that is best for the non-recursive reference
as well.

\begin{table}[t]
\centering
\caption{Distillation loss component weights, evaluated on the recursive
rCLIP~1b12i student. The first block varies the FD weight at fixed CRD and
ICL; the second varies CRD and ICL at an FD weight of $5000$; the third
reports the highest FD weight examined. 
Last block applies the starting loss configuration and final one to the non-recursive model. The selected configuration is shown in bold. Values are top-1 accuracy (\%),
single seed, CC3M, distilled from a DFN-2B ViT-B/16 teacher.}
\setlength{\tabcolsep}{4pt}
\small
\begin{tabular}{lccccccc}
\toprule
Model & CRD & ICL & FD & IN & IN-V2 & IN-R & IN-S \\
\midrule
rCLIP 1b12i  & 1   & 1   & 500    & 26.7 & 23.0 & 28.4 & 13.4 \\
rCLIP 1b12i  & 1   & 1   & 1000   & 27.9 & 24.2 & 30.6 & 14.5 \\
rCLIP 1b12i  & 1   & 1   & 2000   & 30.3 & 26.0 & 33.2 & 16.3 \\
rCLIP 1b12i  & 1   & 1   & 3500   & 31.7 & 27.1 & 35.6 & 18.0 \\
rCLIP 1b12i  & 1   & 1   & 5000   & 32.0 & 28.1 & 34.8 & 17.8 \\
\midrule
rCLIP 1b12i  & 0.5 & 1   & 5000   & 31.9 & 27.9 & 34.9 & 18.2 \\
rCLIP 1b12i  & 2   & 1   & 5000   & 32.5 & 28.3 & 36.1 & 18.6 \\
rCLIP 1b12i  & 1   & 2   & 5000   & 31.3 & 27.1 & 33.6 & 17.4 \\
rCLIP 1b12i  & 2   & 2   & 5000   & 31.3 & 27.5 & 34.7 & 17.8 \\
rCLIP 1b12i  & 1   & 0.5 & 5000   & 32.7 & 28.1 & 36.1 & 18.3 \\
rCLIP 1b12i  & 2   & 0.5 & 5000   & 32.6 & 28.2 & 36.8 & 18.4 \\
rCLIP 1b12i  & 1   & 0   & 5000   & 32.5 & 28.4 & 35.0 & 18.1 \\
\midrule
rCLIP 1b12i  & 0.5 & 0.5 & 10000  & 32.9 & 28.9 & 33.9 & 19.7 \\
rCLIP 1b12i  & 1   & 0.5 & 10000  & 33.6 & 29.3 & 36.3 & 20.0 \\
rCLIP 1b12i  & \textbf{1} & \textbf{0} & \textbf{10000} & \textbf{33.8} & \textbf{29.6} & 35.6 & \textbf{20.2} \\
\midrule
Ref. CLIP & 1 & 1 & 2000  & 37.3 & 32.5 & 46.9 & 27.0 \\
Ref. CLIP & 1 & 0 & 10000 & 47.7 & 41.3 & 59.0 & 38.4 \\
\bottomrule
\end{tabular}

\label{tab:loss-weights}
\end{table}

\subsection{Limitations}

Every configuration was trained once, so differences of a few tenths of a point
should not be over-interpreted. The search was conducted on CC3M and the
selected configuration was transferred to the combined CC3M and CC12M corpus
without re-verification. The selected configuration differs from the sweep anchor in two
respects simultaneously --- ICL removed and FD raised --- and we did not train
the intermediate point (CRD $=1$, ICL $=1$, FD $=10^4$), so the gain cannot be
attributed to either change individually.

Regarding the AM result of Sect.~\ref{sec:loss-choice}, the evidence rests on a single recursive configuration (1b12i) at two weights. Sect.~\ref{app:am-weight}
rules out under-weighting as an explanation, but does not establish how the effect scales with the number of distinct blocks. Our account predicts that the penalty should diminish as $M$ decreases, since fewer teacher states are then assigned to each parameter set; we did not test this, and configurations with
$N > 1$ may behave differently.
The weighting search was conducted on rCLIP~1b12i and the selected
configuration was applied unchanged to the r²CLIP family,
whose text tower is also recursive; we did not re-verify it there.

\section{Extended evaluations}
\label{sec:supp_extended}

This section reports the complete block/iteration grid for the experiments summarized in the main text, where for brevity only the 1b12i, 3b4i and 6b2i configurations were shown. We first give the exact values underlying Fig.~\ref{fig:baseline-clip} and Fig.~\ref{fig:distillation-results} (Sect.~\ref{sec:supp_ext_pretrain}), then the full grids for few-shot classification, visual grounding and unlearning (Sects.~\ref{sec:supp_ext_fewshot}--\ref{sec:supp_ext_unlearning}). The trends have already been discussed in Sect.~\ref{sec:experiments} and are not restated here; we note only where the intermediate configurations add something the reduced grid does not show.

\subsection{Zero-shot classification and retrieval}
\label{sec:supp_ext_pretrain}

Table~\ref{tab:clip_results} and Table~\ref{tab:clip_results_distill_loss} give the exact values plotted in the main text, without and with distillation respectively. The gain from additional distinct blocks is heavily front-loaded: without distillation, rCLIP recovers $2.7$ of its total $3.7$-point ImageNet improvement between one and two blocks, and the $4$- and $6$-block configurations are indistinguishable at $36.5$. Under distillation the curve is less saturated but the same ordering holds, with the step from one to two blocks worth $6.1$ points against $2.1$ for the step from four to six. The two recursive families separate only under distillation. Trained directly with the CLIP objective, r²CLIP matches or slightly exceeds rCLIP at most grid points except the retrieval tasks, despite its smaller text tower. After distillation, rCLIP leads at every configuration across most datasets for both lower and higher block counts.

\begin{table*}[t]
\centering
\caption{Zero-shot classification and cross-modal retrieval across the full block/iteration grid, trained directly with the CLIP objective and no knowledge distillation. Values are top-1 accuracy for classification [\%] and Recall@1 [\%] for retrieval. Ref. CLIP is the non-recursive ViT-B/16 trained under the same recipe. Bold marks the best recursive model per column and underline the best model overall. Exact values underlying Fig.~\ref{fig:baseline-clip} in the main paper.}
\begin{tabular}{lcccc||cccc}
\toprule
\multirow{2}{*}{CLIP Model}  & \multirow{2}{*}{ImageNet} & \multirow{2}{*}{ImageNet-V2} & \multirow{2}{*}{ImageNet-R} & \multirow{2}{*}{ImageNet-S} & \multicolumn{2}{c}{MSCOCO} & \multicolumn{2}{c}{Flickr}  \\
\cline{6-9}
 & & &  & &  I2T & T2I & I2T & T2I \\
\midrule
Ref. CLIP & \underline{38.6} & \underline{33.2} & \underline{49.3} & \underline{26.4} & \underline{36.4} & \underline{26.5} & \underline{63.0} & \underline{48.9} \\
\midrule
rCLIP 1b12i & 32.8 & 28.6 & 36.8 & 17.8 & 30.6 & 22.4 & 57.6 & 42.9  \\
rCLIP 2b6i & 35.5 & 30.5 & 42.4 & 21.0 & 33.9 & 24.4 & 58.3 & 45.9  \\
rCLIP 3b4i & 36.2 & 30.6 & 45.0 & 23.3 & \textbf{35.5} & 25.0 & 60.0 & 47.6  \\
rCLIP 4b3i & 36.5 & 31.1 & 45.6 & 23.4 & 34.4 & 25.0 & 59.6 & 47.8  \\
rCLIP 6b2i & 36.5 & 31.7 & 45.3 & 23.6 & 35.0 & \textbf{25.5} & 60.2 & \textbf{48.6}  \\
\midrule\midrule
r²CLIP 1b12i & 33.4 & 29.0 & 39.2 & 18.6 & 31.9 & 22.1 & 56.4 & 43.4  \\
r²CLIP 2b6i  & 35.5 & 30.1 & 41.7 & 20.5 & 34.6 & 23.5 & 58.5 & 45.4  \\
r²CLIP 3b4i & \textbf{36.6} & \textbf{32.0} & \textbf{46.7} & 23.5 & 35.4 & 24.8 & 60.6 & 46.5  \\
r²CLIP 4b3i  & 36.4 & 31.7 & 45.7 & 22.8 & 34.9 & 24.2 & 60.0 & 46.4  \\
r²CLIP 6b2i  & \textbf{36.6} & 31.6 & \textbf{46.7} & \textbf{23.7} & 35.4 & 25.0 & \textbf{61.6} & 48.2  \\
\bottomrule
\end{tabular}%

\label{tab:clip_results}
\end{table*}

\begin{table*}[tb]
\centering
\caption{Zero-shot classification and cross-modal retrieval across the full block/iteration grid, distilled from the DFN-2B ViT-B/16 teacher on CC3M + CC12M under the objective selected in Sect.~\ref{sec:loss-choice} and Sect. \ref{app:loss-weighting} ($\lambda_{\mathrm{CRD}}=1$, $\lambda_{\mathrm{ICL}}=0$, $\lambda_{\mathrm{FD}}=10^4$). Values are top-1 accuracy for classification [\%] and Recall@1 [\%] for retrieval. Bold marks the best recursive model per column and underline the best model overall. Exact values underlying Fig.~\ref{fig:distillation-results} in the main paper.}
\begin{tabular}{lcccc||cccc}
\toprule
\multirow{2}{*}{CLIP Model}  & \multirow{2}{*}{ImageNet} & \multirow{2}{*}{ImageNet-V2} & \multirow{2}{*}{ImageNet-R} & \multirow{2}{*}{ImageNet-S} & \multicolumn{2}{c}{MSCOCO} & \multicolumn{2}{c}{Flickr}  \\
\cline{6-9}
 & & &  & &  I2T & T2I & I2T & T2I \\
\midrule
Teacher & \underline{76.29} & \underline{68.15} & \underline{83.10} & \underline{62.05} & 60.42 & 43.39 & 85.50 & 69.10 \\
\midrule
Ref. CLIP & 72.50 & 64.30 & 81.60 & 58.50 & \underline{61.04} & \underline{44.00} & \underline{86.20} & \underline{70.00} \\
\midrule
rCLIP 1b12i & 56.70 & 48.80 & 62.60 & 40.80 & 51.48 & 34.68 & 77.10 & 61.10  \\
rCLIP 2b6i & 62.80 & 54.60 & 70.70 & 48.10 & 55.68 & 38.19 & 81.80 & 65.68  \\
rCLIP 3b4i & 65.20 & 57.00 & 73.30 & 50.50 & 57.36 & 39.68 & 83.50 & 66.94  \\
rCLIP 4b3i & 66.70 & 58.60 & 74.80 & 52.10 & 57.74 & 40.29 & 83.50 & 67.28  \\
rCLIP 6b2i & \textbf{68.80} & \textbf{60.60} & 77.20 & \textbf{54.20} & \textbf{58.98} & \textbf{41.83} & \textbf{85.40} & \textbf{68.56}  \\
\midrule\midrule
r²CLIP 1b12i & 52.00 & 44.70 & 61.50 & 37.20 & 47.30 & 32.59 & 72.80 & 58.88  \\
r²CLIP 2b6i & 59.80 & 51.90 & 69.90 & 45.60 & 53.32 & 37.15 & 79.90 & 63.14  \\
r²CLIP 3b4i & 62.70 & 54.80 & 72.00 & 48.30 & 54.34 & 38.54 & 83.10 & 65.58  \\
r²CLIP 4b3i & 65.20 & 56.70 & 74.50 & 51.10 & 56.60 & 39.62 & 83.50 & 66.28  \\
r²CLIP 6b2i & 67.70 & 60.00 & \textbf{77.50} & 53.90 & 58.06 & 41.62 & 83.70 & 68.08  \\
\bottomrule
\end{tabular}%

\label{tab:clip_results_distill_loss}
\end{table*}

\subsection{Few-shot classification}
\label{sec:supp_ext_fewshot}

Table~\ref{tab:fewshot-results-full} gives the full grid for zero-shot and 16-shot classification. Averaged over the ten datasets, the gap to the distilled Reference CLIP narrows monotonically with the number of distinct blocks, from $12.5$ to $3.2$ points for rCLIP and from $16.4$ to $4.5$ for r²CLIP in the zero-shot setting. Few-shot adaptation reduces every one of these gaps, to $10.5$ and $2.7$ points for rCLIP at one and six blocks and to $13.6$ and $3.0$ for r²CLIP, so the benefit of adaptation is largest where the backbone is most constrained.

The intermediate configurations show that the cost of text-tower recursion is essentially removed by adaptation. The average gap between rCLIP and r²CLIP at matched block counts falls from $3.9$ to $1.3$ points zero-shot across the grid, and from $3.0$ to $0.3$ points after few-shot adaptation. At six blocks the two families are indistinguishable under adaptation, and r²CLIP in fact attains the best recursive result on five of the ten datasets.

Two departures from the otherwise monotone trend are worth noting. EuroSAT is non-monotone in block count for both families and in both settings, and is also the dataset with by far the largest adaptation gain --- between $47$ and $56$ points for every recursive backbone, against $46$ for the reference --- which is consistent with a large domain gap from the pretraining distribution rather than with a property of recursion. Aircraft shows the opposite pattern: adaptation gains grow with block count ($30.8$ to $38.9$ points for rCLIP), and it is the only dataset where the recursive models remain far from the reference after adaptation, reflecting how little fine-grained aircraft signal survives distillation on CC3M and CC12M.

\begin{table*}[tb]
\centering
\caption{Zero-shot and 16-shot accuracy [\%] across ten classification datasets, for the full block/iteration grid. Few-shot results use the CLIP-LoRA~\cite{zanella2024low} adaptation method with $16$ shots per class; the configuration is given in Supplementary ~\ref{sec:supp_fewshot_hparams}. Within each setting, bold marks the best recursive model per column and underline the best model overall. Extended version of Table~\ref{tab:fewshot-results}  in the main paper.}
\label{tab:fewshot-results-full}
\resizebox{\textwidth}{!}{%
\begin{tabular}{llcccccccccc}
\toprule
Setting & CLIP Model & ImageNet & SUN & Aircraft & EuroSAT & Food
& Pets & Flowers & Caltech & DTD & UCF \\
\midrule
\multirow{13}{*}{Zero-shot}
& ViT-B/16 (OAI chkpt) & 66.73 & 62.57 & \underline{23.97} & 42.09 & 85.27 & 88.28 & 67.28 & 93.23 & 44.39 & \underline{65.16} \\
& Teacher & \underline{75.91} & \underline{70.68} & 21.60 & \underline{48.09} & \underline{88.27} & 91.47 & \underline{79.09} & \underline{96.96} & \underline{56.86} & 61.83 \\
& Ref. CLIP & 71.86 & 70.45 & 16.59 & 44.43 & 81.05 & \underline{91.66} & 66.10 & 95.46 & 55.67 & 60.19 \\
\cmidrule(lr){2-12}
& rCLIP 1b12i & 56.34 & 61.22 & 4.35 & 33.53 & 58.80 & 84.57 & 48.76 & 89.45 & 42.49 & 49.38 \\
& rCLIP 2b6i & 62.13 & 64.90 & 7.98 & 37.22 & 67.62 & 86.10 & 57.53 & 91.28 & 48.82 & 52.50 \\
& rCLIP 3b4i & 64.65 & 66.30 & 8.73 & \textbf{41.88} & 70.63 & 87.24 & 59.11 & 92.54 & 50.77 & 53.61 \\
& rCLIP 4b3i & 66.06 & 67.75 & 10.59 & 39.73 & 73.20 & 89.07 & 59.24 & 93.67 & 48.94 & 55.19 \\
& rCLIP 6b2i & \textbf{68.54} & 68.22 & \textbf{12.87} & 39.83 & \textbf{75.39} & \textbf{89.81} & \textbf{61.92} & 93.55 & \textbf{54.43} & 56.54 \\
\cmidrule(lr){2-12}
& r²CLIP 1b12i & 51.05 & 59.11 & 4.62 & 32.01 & 53.64 & 80.62 & 35.57 & 88.24 & 37.65 & 47.32 \\
& r²CLIP 2b6i & 58.70 & 63.32 & 8.52 & 37.16 & 63.14 & 82.61 & 48.80 & 90.14 & 42.91 & 50.33 \\
& r²CLIP 3b4i & 62.05 & 65.01 & 8.97 & 33.99 & 68.41 & 87.33 & 54.61 & 92.21 & 46.04 & 53.32 \\
& r²CLIP 4b3i & 64.28 & 66.59 & 10.47 & 37.42 & 70.66 & 87.65 & 55.30 & 92.78 & 48.11 & 54.69 \\
& r²CLIP 6b2i & 66.93 & \textbf{68.40} & 12.66 & 36.81 & 75.10 & 86.13 & 59.64 & \textbf{94.04} & 51.48 & \textbf{56.89} \\
\midrule\midrule
\multirow{13}{*}{Few-shot (16)}
& ViT-B/16 (OAI chkpt) & 73.37 & 76.22 & 55.54 & 90.65 & 84.74 & 91.96 & 98.01 & 96.39 & 71.99 & \underline{86.23} \\
& Teacher & \underline{77.83} & 76.92 & \underline{61.60} & \underline{91.58} & \underline{87.11} & 92.53 & \underline{99.27} & \underline{97.53} & 75.83 & 83.11 \\
& Ref. CLIP & 75.18 & \underline{77.21} & 58.78 & 90.74 & 84.58 & \underline{92.78} & 99.03 & 97.24 & \underline{75.95} & 82.42 \\
\cmidrule(lr){2-12}
& rCLIP 1b12i & 61.45 & 69.43 & 35.13 & 87.09 & 67.61 & 85.47 & 94.07 & 94.08 & 61.29 & 72.98 \\
& rCLIP 2b6i & 66.77 & 72.48 & 42.87 & 87.53 & 74.19 & 89.13 & 96.47 & 95.46 & 65.72 & 76.18 \\
& rCLIP 3b4i & 68.85 & 73.91 & 44.55 & 88.58 & 76.98 & 90.13 & 97.16 & 95.94 & 70.98 & 77.95 \\
& rCLIP 4b3i & 70.03 & 74.39 & 47.04 & 88.57 & 78.76 & 91.25 & 97.12 & 96.47 & 69.09 & 79.94 \\
& rCLIP 6b2i & \textbf{72.13} & 75.14 & \textbf{51.79} & 89.36 & 80.18 & 91.33 & \textbf{98.09} & \textbf{96.63} & 71.10 & \textbf{81.13} \\
\cmidrule(lr){2-12}
& r²CLIP 1b12i & 55.35 & 66.05 & 26.04 & 87.54 & 65.27 & 87.30 & 87.37 & 92.78 & 61.35 & 69.13 \\
& r²CLIP 2b6i & 63.37 & 71.46 & 33.96 & 88.00 & 73.15 & 90.46 & 93.26 & 94.69 & 66.08 & 75.36 \\
& r²CLIP 3b4i & 66.48 & 72.91 & 41.01 & 87.07 & 75.83 & 90.22 & 95.21 & 95.25 & 67.85 & 76.63 \\
& r²CLIP 4b3i & 68.62 & 74.45 & 44.04 & 88.73 & 77.84 & 92.01 & 96.63 & 95.78 & 69.62 & 78.06 \\
& r²CLIP 6b2i & 71.44 & \textbf{75.69} & 48.30 & \textbf{89.70} & \textbf{80.44} & \textbf{92.59} & 97.08 & 96.23 & \textbf{72.99} & 79.36 \\
\bottomrule
\end{tabular}%
}
\end{table*}

\subsection{Visual grounding}
\label{sec:supp_ext_grounding}

Table~\ref{tab:grounding-results-full} gives the full grid for grounding accuracy. Averaged over the ten splits, the spread across the entire block/iteration range is only $3.4$ points for rCLIP ($71.9$ at one block to $75.3$ at six) and $3.5$ for r²CLIP ($72.1$ to $75.6$), against $75.7$--$76.6$ for the three non-recursive references. Even the most constrained configuration therefore loses less than five points on average relative to a full-depth model, and the best recursive configurations come within $0.2$ points of the strongest reference on RefCOCO testB and RefCOCO+ val, and exceed it on ReferItGame.

The intermediate configurations confirm that neither family dominates. Averaged over the ten splits, r²CLIP is marginally ahead of rCLIP at four of the five block counts, by between $0.1$ and $0.5$ points, with the two families within $0.6$ points everywhere; the best recursive result is attained by an r²CLIP configuration on six of the ten splits and by an rCLIP configuration on four. This is in clear contrast to classification and retrieval, where rCLIP leads at every matched configuration, and supports the reading in Sect.~\ref{sec:downstream-tasks} that HiVG grounding is largely insensitive to whether recursion is applied to one encoder or both.

\begin{table*}[t]
\centering
\caption{Visual grounding accuracy [\%] at IoU $\geq 0.5$ across the full block/iteration grid, following the HiVG~\cite{xiao2024hivg} adaptation protocol; the configuration is given in Supplementary~\ref{sec:supp_ground_hparams}. Teacher denotes the DFN-2B ViT-B/16 and Ref. CLIP the non-recursive ViT-B/16 distilled from it under the same recipe. Unlike HiVG we omit LoRA pre-adaptation on MSCOCO and adapt each checkpoint directly to the target dataset, so absolute values are not directly comparable to the published ones. Bold marks the best recursive model per column and underline the best model overall. Extended version of Table~\ref{tab:grounding-results}  in the main paper.}
\label{tab:grounding-results-full}
\begin{tabular}{l|ccc|ccc|cc|c|c}
\toprule
\multirow{2}{*}{\makecell{CLIP Model}} & \multicolumn{3}{c|}{RefCOCO} &
\multicolumn{3}{c|}{RefCOCO+} & \multicolumn{2}{c|}{RefCOCOg} & ReferIt & Flickr \\
& val & testA & testB & val & testA & testB & val & test & test & test \\
\midrule
ViT-B/16 (OAI chkpt) & 83.99 & \underline{86.86} & 78.45 & \underline{73.96} & \underline{81.98} & 63.06 & 74.16 & \underline{74.39} & 68.62 & \underline{80.85} \\
Teacher & 83.03 & 85.97 & 78.26 & 72.89 & 79.19 & 63.10 & 73.43 & 73.52 & 68.26 & 79.65 \\
Ref. CLIP & \underline{84.31} & 86.68 & \underline{78.85} & 73.21 & 78.46 & \underline{64.74} & \underline{75.25} & 74.00 & 68.98 & 80.18 \\
\midrule
rCLIP 1b12i & 80.08 & 82.81 & 74.55 & 67.61 & 74.13 & 57.95 & 69.26 & 68.73 & 67.12 & 76.44 \\
rCLIP 2b6i & 81.66 & 83.34 & 76.57 & 70.07 & 75.44 & 60.85 & 71.22 & 70.82 & 67.74 & 78.16 \\
rCLIP 3b4i & 81.75 & 84.56 & 76.96 & 70.99 & 76.78 & 61.30 & 72.24 & 71.93 & 68.19 & 78.28 \\
rCLIP 4b3i & 82.13 & 85.13 & 76.94 & 71.97 & 77.27 & 63.45 & 71.73 & 72.00 & 68.98 & 78.62 \\
rCLIP 6b2i & \textbf{83.24} & 85.05 & \textbf{78.73} & 72.38 & 77.23 & 63.45 & 71.94 & 71.64 & \textbf{\underline{69.41}} & \textbf{79.40} \\
\midrule
r²CLIP 1b12i & 79.95 & 82.70 & 75.16 & 68.08 & 75.31 & 59.77 & 68.42 & 68.57 & 66.58 & 76.36 \\
r²CLIP 2b6i & 81.72 & 84.77 & 77.08 & 70.78 & 75.86 & 61.30 & 71.18 & 71.73 & 68.03 & 78.53 \\
r²CLIP 3b4i & 82.04 & 85.05 & 76.22 & 71.06 & 77.32 & 62.57 & 72.10 & 71.27 & 68.24 & 78.53 \\
r²CLIP 4b3i & 82.85 & \textbf{85.51} & \textbf{78.04} & 70.96 & 76.38 & 61.14 & 72.51 & \textbf{72.27} & 69.11 & 78.58 \\
r²CLIP 6b2i & 82.80 & 84.88 & \textbf{78.04} & \textbf{73.90} & \textbf{78.95} & \textbf{64.10} & \textbf{72.94} & 71.78 & \textbf{\underline{69.41}} & 79.27 \\
\bottomrule
\end{tabular}
\end{table*}

\subsection{Concept unlearning}
\label{sec:supp_ext_unlearning}

Table~\ref{tab:diet_results_full} gives the full grid for zero-shot and post-DIET accuracy. Because the backbones start from very different zero-shot accuracies, absolute drops in $D_t$ are not comparable across models; we therefore also read retention as the ratio $D_t /  D_{t0}$ of average test accuracy after unlearning to before. On this measure the non-recursive references retain $82.8\%$ (OpenAI), $80.5\%$ (Teacher) and $72.2\%$ (Reference CLIP), against $57.6$--$74.4\%$ across the recursive grid. Forgetting itself is effective everywhere: average $D_f$ falls from $73.0$--$88.7\%$ before unlearning to $4.0$--$15.2\%$ after, with no systematic difference between recursive and non-recursive backbones.

The intermediate configurations make the trend reported in Sect.~\ref{sec:downstream-tasks} look less regular than the reduced grid suggests. Retention broadly improves with the number of distinct blocks for r²CLIP ($66.8$, $68.7$, $72.3$, $70.5$ and $74.4\%$ from one to six blocks), but for rCLIP the two most constrained configurations fall well below the rest, at $59.8\%$ and $57.6\%$ against $71.1$--$71.5\%$ for the three larger ones. The rCLIP 2b6i is the most extreme: it attains the lowest average post-DIET $D_f$ of any model in the grid ($4.0\%$) together with the second-lowest retention, and its per-dataset test accuracies carry standard deviations of up to $17.8$ points over five seeds. We attribute this to the sensitivity of DIET to learning-rate selection rather than to the architecture. Our selection criterion is a harmonic mean of forgetting efficacy and retention, and efficacy saturates once forgetting is close to complete; where no gentler grid point forgets adequately, the selected arm can therefore already have sacrificed substantial retention. Individual configurations should accordingly not be read as characterizing their block count, and the reliable observation is the weaker one that heavily shared backbones retain less on average than the non-recursive references.

\begin{table*}[t]
\centering
\caption{Zero-shot and post-DIET results across the full block/iteration grid. $D_f$ is forget-set and $D_t$ test-set accuracy [\%], averaged over five seeds with one randomly selected class removed per seed; lower $D_f$ indicates more effective forgetting and higher $D_t$ better retention. The Average columns report the unweighted mean over the three datasets. Unlearning uses DIET~\cite{kunananthaseelan2026diet} with a learning rate selected per backbone--dataset pair; the configuration is given in Supplementary~\ref{sec:supp_diet_hparams}. In the post-DIET block, bold marks the best recursive model and underline the best model overall on average $D_t$. Extended version of Table~\ref{tab:diet_results} in the main paper.}
\label{tab:diet_results_full}
\resizebox{\textwidth}{!}{
\begin{tabular}{l cc cc cc cc}
\toprule
\multirow{2}{*}{\textbf{CLIP Model}}
& \multicolumn{2}{c}{\textbf{Flowers102}}
& \multicolumn{2}{c}{\textbf{Pets}}
& \multicolumn{2}{c}{\textbf{Food101}}
& \multicolumn{2}{c}{\textbf{Average}} \\
\cmidrule(lr){2-3}
\cmidrule(lr){4-5}
\cmidrule(lr){6-7}
\cmidrule(lr){8-9}
& $D_f$ (\%) & $D_t$ (\%)
& $D_f$ (\%) & $D_t$ (\%)
& $D_f$ (\%) & $D_t$ (\%)
& $D_f$ (\%) & $D_t$ (\%) \\
\midrule

\multicolumn{9}{l}{\textbf{Zero-shot}} \\

ViT-B/16 (OAI chkpt)
& $92.61{\pm}8.30$ & $70.56{\pm}0.12$
& $83.25{\pm}21.55$ & $89.12{\pm}0.37$
& $87.68{\pm}4.34$ & $85.85{\pm}0.05$
& $87.85$ & $81.84$ \\

Teacher
& $83.50{\pm}27.84$ & $80.35{\pm}0.16$
& $92.00{\pm}9.30$ & $93.34{\pm}0.23$
& $90.24{\pm}3.13$ & $87.77{\pm}0.05$
& $88.58$ & $87.15$ \\

Ref. CLIP
& $88.39{\pm}19.50$ & $66.72{\pm}0.13$
& $89.25{\pm}13.91$ & $92.18{\pm}0.29$
& $88.44{\pm}2.75$ & $80.23{\pm}0.04$
& $88.69$ & $79.71$ \\

\midrule

rCLIP 1b12i
& $66.96{\pm}14.31$ & $49.22{\pm}0.25$
& $85.49{\pm}11.87$ & $87.20{\pm}0.22$
& $66.56{\pm}11.19$ & $58.37{\pm}0.10$
& $73.00$ & $64.93$ \\

rCLIP 2b6i
& $73.20{\pm}21.11$ & $58.74{\pm}0.16$
& $76.75{\pm}18.04$ & $87.93{\pm}0.22$
& $75.08{\pm}7.17$ & $67.28{\pm}0.08$
& $75.01$ & $71.32$ \\

rCLIP 3b4i
& $75.00{\pm}28.97$ & $60.70{\pm}0.13$
& $83.00{\pm}13.27$ & $88.66{\pm}0.26$
& $79.04{\pm}8.04$ & $71.05{\pm}0.07$
& $79.01$ & $73.47$ \\

rCLIP 4b3i
& $80.19{\pm}26.38$ & $62.02{\pm}0.16$
& $85.50{\pm}12.89$ & $90.17{\pm}0.19$
& $80.20{\pm}8.33$ & $72.38{\pm}0.08$
& $81.96$ & $74.86$ \\

rCLIP 6b2i
& $81.51{\pm}26.65$ & $63.44{\pm}0.17$
& $84.00{\pm}15.85$ & $90.65{\pm}0.25$
& $82.84{\pm}6.36$ & $74.78{\pm}0.05$
& $82.78$ & $76.29$ \\

\midrule

r²CLIP 1b12i
& $68.85{\pm}21.92$ & $36.40{\pm}0.31$
& $78.98{\pm}18.49$ & $83.42{\pm}0.42$
& $71.24{\pm}4.53$ & $54.15{\pm}0.04$
& $73.02$ & $57.99$ \\

r²CLIP 2b6i
& $81.89{\pm}16.39$ & $51.25{\pm}0.21$
& $82.99{\pm}17.55$ & $85.71{\pm}0.32$
& $74.72{\pm}4.58$ & $63.33{\pm}0.04$
& $79.87$ & $66.76$ \\

r²CLIP 3b4i
& $88.70{\pm}7.79$ & $57.14{\pm}0.13$
& $85.00{\pm}15.08$ & $89.26{\pm}0.19$
& $76.24{\pm}6.20$ & $68.43{\pm}0.06$
& $83.31$ & $71.61$ \\

r²CLIP 4b3i
& $87.71{\pm}12.12$ & $54.75{\pm}0.16$
& $83.00{\pm}17.08$ & $89.41{\pm}0.25$
& $81.04{\pm}4.57$ & $70.53{\pm}0.05$
& $83.92$ & $71.56$ \\

r²CLIP 6b2i
& $81.88{\pm}24.64$ & $59.06{\pm}0.18$
& $84.50{\pm}14.96$ & $89.48{\pm}0.26$
& $79.76{\pm}8.21$ & $73.05{\pm}0.08$
& $82.05$ & $73.86$ \\

\midrule\midrule

\multicolumn{9}{l}{\textbf{After DIET}} \\

ViT-B/16 (OAI chkpt)
& $4.36{\pm}5.06$ & $56.60{\pm}2.06$
& $2.00{\pm}2.59$ & $73.79{\pm}9.93$
& $12.72{\pm}14.34$ & $72.80{\pm}4.00$
& $6.36$ & $67.73$ \\

Teacher
& $3.87{\pm}6.50$ & $74.24{\pm}3.46$
& $5.01{\pm}4.68$ & $77.18{\pm}4.75$
& $13.08{\pm}11.63$ & $59.15{\pm}11.14$
& $7.32$ & $\underline{70.19}$ \\

Ref. CLIP
& $2.60{\pm}3.16$ & $56.45{\pm}4.57$
& $3.00{\pm}5.35$ & $69.10{\pm}14.18$
& $14.12{\pm}12.58$ & $47.10{\pm}13.61$
& $6.57$ & $57.55$ \\

\midrule

rCLIP 1b12i
& $4.29{\pm}6.27$ & $33.48{\pm}10.18$
& $12.58{\pm}12.10$ & $45.62{\pm}9.19$
& $9.32{\pm}9.26$ & $37.46{\pm}8.65$
& $8.73$ & $38.85$ \\

rCLIP 2b6i
& $1.54{\pm}3.44$ & $31.87{\pm}7.97$
& $5.79{\pm}6.54$ & $57.35{\pm}14.05$
& $4.64{\pm}1.39$ & $34.10{\pm}17.80$
& $3.99$ & $41.11$ \\

rCLIP 3b4i
& $7.20{\pm}8.16$ & $51.92{\pm}2.83$
& $22.72{\pm}36.72$ & $59.82{\pm}10.85$
& $15.52{\pm}6.62$ & $44.93{\pm}4.72$
& $15.15$ & $52.22$ \\

rCLIP 4b3i
& $2.38{\pm}4.16$ & $46.86{\pm}9.74$
& $5.53{\pm}4.43$ & $59.98{\pm}11.75$
& $17.32{\pm}4.62$ & $53.43{\pm}11.78$
& $8.41$ & $53.42$ \\

rCLIP 6b2i
& $1.84{\pm}2.06$ & $44.67{\pm}8.61$
& $4.02{\pm}2.44$ & $67.68{\pm}6.17$
& $9.56{\pm}5.12$ & $51.19{\pm}8.84$
& $5.14$ & $54.51$ \\

\midrule

r²CLIP 1b12i
& $5.67{\pm}5.45$ & $27.73{\pm}4.16$
& $13.14{\pm}24.61$ & $56.63{\pm}13.02$
& $5.96{\pm}7.42$ & $31.77{\pm}3.09$
& $8.26$ & $38.71$ \\

r²CLIP 2b6i
& $5.36{\pm}7.84$ & $39.58{\pm}4.96$
& $2.01{\pm}1.92$ & $52.31{\pm}10.92$
& $16.40{\pm}6.69$ & $45.78{\pm}6.05$
& $7.92$ & $45.89$ \\

r²CLIP 3b4i
& $5.15{\pm}8.67$ & $40.12{\pm}11.39$
& $13.03{\pm}12.00$ & $64.77{\pm}9.37$
& $9.40{\pm}7.32$ & $50.39{\pm}4.73$
& $9.19$ & $51.76$ \\

r²CLIP 4b3i
& $2.83{\pm}1.89$ & $42.50{\pm}6.69$
& $10.52{\pm}9.62$ & $58.25{\pm}12.22$
& $9.48{\pm}3.56$ & $50.67{\pm}10.52$
& $7.61$ & $50.47$ \\

r²CLIP 6b2i
& $0.82{\pm}1.13$ & $44.30{\pm}11.42$
& $6.27{\pm}6.19$ & $68.28{\pm}11.80$
& $9.56{\pm}5.53$ & $52.27{\pm}7.47$
& $5.55$ & $\mathbf{54.95}$ \\

\bottomrule
\end{tabular}
}
\end{table*}

\section{Hyperparameters}
In this section we provide extended information about the experimental setup. All experiments were run on 1 or 4 GPUs (Nvidia H100 96GB or Nvidia RTX 6000 96GB Blackwell Pro).

\subsection{CLIP and distillation training configuration}
All main baseline CLIP and distillation runs share a single recipe, only the trained architecture varies between them. The models are optimized with AdamW using a maximum learning rate of $1\times10^{-3}$, $\beta_1 = 0.9$, $\beta_2 = 0.98$, and $\epsilon = 1\times10^{-10}$. The weight decay is set to $0.1$ and applied to the weight matrices, the projection and loss parameters only; biases, normalization parameters, and the learnable logit scale are excluded from decay. Gradients are clipped to a global $\ell_2$ norm of $1.0$, and the learnable logit scale is clamped to at most $\ln 100$ throughout training. The learning rate follows a linear warmup over $10{,}000$ optimizer steps, approximately $0.74$ of an epoch, and then decays on a cosine schedule; warmup is specified as a fixed number of steps rather than as a fraction of training, so it is identical across all runs.

Training uses the combined CC3M and CC12M datasets, comprising $13{,}874{,}493$ image--text pairs, for $32$ epochs. The effective batch size is fixed at $1024$ across runs, with micro-batch of $256$ across 4 GPUs; this gives $13{,}549$ steps per epoch and $433{,}568$ steps in total, or roughly $444$M samples seen. Text is tokenized to a context length of $77$ tokens.

During training, images are augmented with a random resized crop covering $90$--$100\%$ of the source image area at an aspect ratio sampled from $[3/4, 4/3]$, then resized to $224\times224$; no horizontal flip, color jitter or grayscale augmentation is applied. At evaluation, images are resized to $224$ on the short side and center-cropped to $224\times224$. Both paths use bicubic interpolation with antialiasing and normalize with standard CLIP statistics, mean $(0.4815, 0.4578, 0.4082)$ and standard deviation $(0.2686, 0.2613, 0.2758)$.

Runs are executed with distributed data parallelism in BF16 mixed precision, with an evaluation batch size of $256$ and a fixed random seed of $42$.

\subsection{Visual grounding configuration}
\label{sec:supp_ground_hparams}

All grounding runs share a single recipe following HiVG~\cite{xiao2024hivg}; only the vision and text backbones vary between them, and every backbone is fine-tuned separately on each dataset. The pre-trained encoders remain frozen: gradients reach only the LoRA adapters, the cross-modal bridges, the fusion transformer and the prediction heads. All trainable parameters live in a single optimizer group, so there is no separate backbone or head learning rate.
\paragraph{Optimization.}
We use AdamW with $\beta_1 = 0.9$, $\beta_2 = 0.999$, $\epsilon = 1\times10^{-8}$ and weight decay $1\times10^{-4}$, with gradients clipped to a global $\ell_2$ norm of $1.0$. Training proceeds in curriculum phases (Table~\ref{tab:ground_curriculum}); each phase resumes from the previous phase's best-validation checkpoint and restarts a cosine schedule from its own peak learning rate, stepped once per epoch and without warmup. Peak rates fall from $2.5\times10^{-4}$ in the warmup phase to $2.5\times10^{-5}$ in the final stage. The effective batch size is fixed per phase across every backbone at four times the per-GPU value, running from $640$ down to $320$ over the curriculum, so the optimization trajectory is identical regardless of the backbone; incomplete batches are dropped during training only. Runs use distributed data parallelism in FP32 with seed $13$ offset by process rank; we select the checkpoint with the best validation Acc@$0.5$ and report top-1 accuracy at an IoU threshold of $0.5$.

\paragraph{Curriculum.}
The Hi-LoRA curriculum consists of $1 + \min(N, 3)$ phases, where $N$ is the number of \emph{distinct} transformer blocks in the backbone. In the warmup phase the backbone carries no adapters and only the bridges, fusion transformer and heads are trained; each subsequent stage unlocks vision adapters on a further fraction of the distinct blocks, cumulatively. Because adapters are tied to unique weight matrices rather than to depth positions, a backbone with fewer than three distinct blocks cannot unlock a new one at every stage and therefore runs a correspondingly shorter curriculum; for the non-recursive twelve-block backbones the cumulative schedule recovers HiVG's original $5/8/12$ unlock pattern. The text-tower adapters are unlocked flat, across all blocks at once, from stage~1 onward.
Because adapters are tied to distinct blocks rather than to depth
positions, blocks unlocked in early stages are re-executed throughout
the unrolled encoder: for $N = 4$, $M = 3$, adapting the first block
adapts execution steps $1$, $5$ and $9$. The curriculum therefore
stages adaptation over parameter groups rather than over network
depth, and coincides with HiVG's shallow-to-deep schedule only in the
non-recursive case.

\begin{table}[t]
\centering
\footnotesize
\setlength{\tabcolsep}{3.5pt}
\caption{Hi-LoRA curriculum. Each phase resumes from the previous
phase's best-validation checkpoint and restarts the cosine schedule.
The number of phases is $1 + \min(N, 3)$, where $N$ is the number of
distinct transformer blocks: adapters are tied to unique blocks, so a
backbone with $N < 3$ cannot unlock a new one at every stage. Vision
adapters are unlocked cumulatively over three groups of distinct
blocks, split as evenly as possible with any remainder assigned to the
earlier groups; for $N = 12$ this gives $5/8/12$, reproducing HiVG's
original unlock pattern. The \emph{Blocks} column refers to the vision
tower only: text-tower adapters are unlocked across all blocks at once
from stage~1 onward.}
\label{tab:ground_curriculum}
\begin{tabular}{lcccc}
\toprule
Phase & Blocks$^{\dagger}$ & Epochs$^{*}$ & LR & Batch \\
\midrule
Warmup   & --        & 60\,/\,30 & $2.5\!\times\!10^{-4}$ & 640 \\
Stage 1  & 1st group & 20\,/\,5  & $1.0\!\times\!10^{-4}$ & 480 \\
Stage 2  & 2nd group & 20\,/\,5  & $5.0\!\times\!10^{-5}$ & 480 \\
Stage 3  & all $N$   & 20\,/\,5  & $2.5\!\times\!10^{-5}$ & 320 \\
\bottomrule
\end{tabular}

\vspace{2pt}
{\scriptsize
$^{\dagger}$Cumulative unlocks by $N$: $1\!\rightarrow\!1$;
$2\!\rightarrow\!1,2$; $3\!\rightarrow\!1,2,3$; $4\!\rightarrow\!2,3,4$;
$6\!\rightarrow\!2,4,6$; $12\!\rightarrow\!5,8,12$.\\
$^{*}$RefCOCO\,/\,Flickr. RefCOCO, RefCOCO+, RefCOCOg (umd) and
ReferItGame share the $60/20/20/20$ budget; Flickr30k Entities uses
$30/5/5/5$. Batch sizes are effective (per-GPU $\times$ 4).}
\end{table}

\paragraph{Adaptation and fusion.}
LoRA adapters use rank $32$ with $\alpha = 16$, giving a scale of $0.5$, and dropout $0.1$. In the vision tower they are attached to the attention projections and the MLP. Four multi-modal adaptive cross-modal bridges connect the vision tower to the fusion module; for recursive backbones we attach them to \emph{executions} rather than to layers, placing them after executions $1$, $4$, $8$ and $12$ of the unrolled encoder, which for a non-recursive backbone coincides with HiVG's original placement. Each bridge uses a single attention head at width $512$ with dropout $0.1$, and its output projection is initialized to zero so that the bridge begins as a no-op. The number of text-encoder depths concatenated into the bridges is dataset-specific: the final depth alone for RefCOCO and Flickr30k Entities, depths $\{6, 12\}$ for RefCOCO+ and ReferItGame, and all twelve for RefCOCOg. Every hidden state is layer-normalized before concatenation, and since this choice changes the bridge parameter shapes it is held fixed across all phases of a curriculum. The fusion module is a six-layer pre-norm vision--language encoder of width $512$ with eight heads, feed-forward dimension $2048$ and dropout $0.1$, operating on a sequence of $196$ visual tokens plus \texttt{[CLS]}, $77$ text tokens, and a learned \texttt{[REG]} token from which the box is predicted.

\paragraph{Data and augmentation.}
We train on RefCOCO, RefCOCO+, RefCOCOg (umd split), ReferItGame and
Flickr30k Entities in the fully supervised single-dataset setting, at
$224\times224$ resolution with patch size $16$ and text context length $77$.
Every training image is resized so that its long side is drawn uniformly from
$\{224, 192, 160, 128, 96, 64, 32\}$.\footnote{TransVG
 32i schedule. At the original
$640$~px input it spans $640$–$448$; at $224$ the smallest scale reduces the
long side to $32$~px, i.e.\ $2\times2$ patches within the padded canvas. We
keep the published rule.} Before this resize, and only
for expressions containing none of the words \emph{left}, \emph{right},
\emph{top}, \emph{bottom} or \emph{middle}, the image passes with probability
$0.5$ through a short-side resize to $\{400, 500, 600\}$ followed by a
random-size crop; the crop is retried up to twenty times for a window that
retains the target box and is skipped if none is found. Expressions carrying a
directional word always take the plain-resize path, for the same reason that
horizontal flipping is disabled throughout: both transformations invalidate
spatial referring expressions. With probability $0.8$, brightness, contrast and
saturation are then rescaled by independent factors drawn from
$\mathcal{U}(0.6, 1.4)$ and applied in random order; there is no hue jitter and
no Gaussian blur. Images are normalized with ImageNet statistics, mean
$(0.485, 0.456, 0.406)$ and standard deviation
$(0.229, 0.224, 0.225)$,\footnote{These are ImageNet rather than CLIP
normalization statistics, inherited from the HiVG/TransVG data pipeline and
left unchanged so that recursive and non-recursive backbones are adapted under
an identical protocol.} and padded to the square input with a mask marking the
padded region, placed at a random offset during training and centred at
evaluation. Evaluation applies no augmentation beyond the long-side resize to
$224$.

\paragraph{Objective.}
The training objective combines $\ell_1$ and GIoU box losses at weight $2.0$
each, an image--text contrastive term at weight $1.0$, and token-alignment and
segmentation terms each formed as focal loss at weight $20.0$ plus dice loss at
weight $2.0$, with focal parameters $\alpha = 0.25$ and $\gamma = 2$.

\subsection{Few-shot adaptation configuration}
\label{sec:supp_fewshot_hparams}

Few-shot results follow CLIP-LoRA~\cite{zanella2024low} with its published recipe inherited unchanged, so that our distilled backbones and the non-recursive controls are directly comparable. A single configuration is shared by every run: all backbone configurations across all ten datasets use identical settings, with no per-dataset or per-configuration tuning. Only the backbone under adaptation varies.

\paragraph{Adapters.}
LoRA adapters of rank $r = 2$ are inserted into the query, key and value projections of both the vision and the text tower, with $\alpha = 1$ under CLIP-LoRA's scaling convention $\alpha/\sqrt{r}$ rather than the $\alpha/r$ of the original LoRA formulation. The $A$ matrices are initialized with Kaiming-uniform sampling and the $B$ matrices with zeros, so adaptation begins as a no-op, and dropout of $0.25$ is applied at the adapter input. The backbone is frozen throughout, including the logit scale, which is held at $100$ rather than learned. Adapters are attached to every \emph{distinct} block of the recursive core and shared across all iterations of that block, exactly as the frozen weights are; for the non-recursive controls this reduces to one adapter per layer across all twelve vision and twelve text blocks.

This is where the parameter efficiency of the recursive backbone is most visible. With $r = 2$, each adapted $d \times d$ projection adds $2rd$ parameters, so the adapter budget is set by the number of distinct blocks $N$ rather than by depth. The non-recursive control trains $184{,}320$ adapter parameters; rCLIP, whose text tower is not recursive, trains $9{,}216N + 73{,}728$, from $82{,}944$ at $N = 1$ to $129{,}024$ at $N = 6$; and r²CLIP trains $15{,}360N$, from $15{,}360$ at $N = 1$ to $92{,}160$ at $N = 6$. In every case this is at most $0.13\%$ of the backbone parameters, and the most constrained r²CLIP configuration requires an order of magnitude fewer adapter parameters than the control.

\paragraph{Optimization.}
Adapters are trained with AdamW at a learning rate of $2\times10^{-4}$ with $\beta_1 = 0.9$, $\beta_2 = 0.999$, $\epsilon = 1\times10^{-8}$ and weight decay $1\times10^{-2}$ applied uniformly to all LoRA parameters. The learning rate follows cosine annealing to a floor of $1\times10^{-6}$ without warmup, and no gradient clipping is used. Following CLIP-LoRA, the iteration budget is fixed at $500 \times \text{shots}$, giving $8{,}000$ iterations at batch size $32$, or $256{,}000$ samples seen. The number of epochs is therefore a consequence of the support-set size $16C$ for $C$ classes rather than a hyperparameter, ranging from $16$ epochs on ImageNet to $1{,}600$ on EuroSAT. Each run uses a single GPU in FP16 mixed precision.

\paragraph{Data and protocol.}
We use $16$ shots per class on all ten datasets, with the support and test splits of Zhou et al. \cite{zhou2022coop}\ and the official $50{,}000$-image validation set for ImageNet. Training images are augmented with a random resized crop covering $8$--$100\%$ of the source area, resized to $224\times224$, and horizontally flipped with probability $0.5$; no color jitter or grayscale augmentation is used. Evaluation resizes to $224$ and center-crops to $224\times224$. Both paths use bicubic interpolation and the standard CLIP normalization statistics, mean $(0.4815, 0.4578, 0.4082)$ and standard deviation $(0.2686, 0.2613, 0.2758)$. Text is tokenized to a context length of $77$.

Classifiers are built from the single generic template \texttt{"a photo of a \{\}."}, with underscores in class names replaced by spaces, for both the zero-shot reference and the LoRA-tuned classifier. This is CLIP-LoRA's protocol and differs from the $80$-template ensemble used for the zero-shot results of Sect.~\ref{sec:distillation}; the zero-shot column of Table~\ref{tab:fewshot-results} is therefore evaluated under the same protocol as its adjacent few-shot column, and comparisons should be made within a table rather than across. Following the original protocol we perform no validation-based model selection and report final-iterate top-1 accuracy from a single run with seed $1$.

\subsection{Unlearning configuration}
\label{sec:supp_diet_hparams}

Unlearning follows DIET~\cite{kunananthaseelan2026diet} with its published recipe; only the backbone and the unlearning learning rate vary between runs. Each run removes a single forget concept from a frozen CLIP-style backbone using \emph{only} the 16-shot forget set, with no retain data loaded at any point. At batch size $32$ a 16-shot forget set fits in a single batch, so one epoch is one optimizer step and the entire unlearning budget is $30$ steps.

\paragraph{Objective.}
The loss combines three terms, $\lambda_{\mathrm{hyp}}\mathcal{L}_{\mathrm{Buse}} + \lambda_{\mathrm{ot}}\langle\Pi, C\rangle + \lambda_{\mathrm{ret}}\mathcal{L}_{\mathrm{ret}}$, weighted at $30$, $1$ and $20$ respectively. Image features are norm-clipped at radius $1.25$ and mapped into a Poincaré ball of curvature $c = 1$ via the exponential map at the origin, giving a realized feature norm of approximately $0.848$. Prototypes are the text embeddings of the $C - 1$ retained classes, placed radially at the ball boundary with $\|\xi\| = 1 - 10^{-6}$. Transport cost between features and prototypes is the Busemann function $0.95\left[\log\|z - \xi\|^2 - 0.05\log(1 - \|z\|^2)\right]$; the Busemann penalty itself is instantiated with dimension $0$, so it reduces to $2\log\|\xi - z\| - \log(1 - \|z\|^2 + 10^{-6})$ with no dimension-dependent regularizer. Assignments are obtained by entropic Sinkhorn with $\varepsilon = 0.5$ and at most $2{,}000$ iterations under uniform marginals, and each feature is assigned the prototype given by the row-argmax of the resulting plan. The retain term is a retain-data-free surrogate: it penalizes $\mathrm{mean}(0.85 - S_{ij})$ over the in-batch image--image cosine Gram matrix of the forget batch, encouraging the model to preserve similarity structure without access to retained examples.

\paragraph{Adapters and optimization.}
LoRA adapters of rank $4$ with $\alpha = 1$ (effective scale $\alpha/\sqrt{r} = 0.5$) and dropout $0.1$ are attached to the query, key and value projections of the vision tower only; the feedforward and output projections are left unadapted. As elsewhere in this work, adapters are attached to every \emph{distinct} block and shared across iterations, so the trainable count scales with the number of distinct blocks rather than with depth, ranging from $0.02$\,M at a single block to $0.22$\,M for a standard twelve block ViT-B/16. Training uses AdamW with $\beta_1 = 0.9$, $\beta_2 = 0.999$, $\epsilon = 1\times10^{-8}$ and weight decay $1\times10^{-5}$ on all LoRA matrices, under cosine annealing over the $30$ steps to a floor of $1\times10^{-5}$ with no warmup and no gradient clipping. Following the original method, the learning rate is dropped by a factor of ten whenever forget-set validation accuracy falls below $2\%$, measured on at most eight held-out images of the forget class, with a two-step cooldown between drops.

\paragraph{Learning-rate selection.}
DIET's published global learning rate is mis-tuned for backbones with wider forget-class margins, so we perform a search for every evaluated model over $0.0009 \cdot 2^{n}$ for $n \in \{-2, -1, 0, 1, 2, 3, 4\}$, bracketing the published rate on both sides, and apply the same grid to the reference backbones. For each backbone--dataset pair we select the rate maximizing the harmonic mean of a forgetting effectiveness score
\begin{equation}
e = \mathrm{clip}\!\left(\frac{D_{f0} - D_f}{D_{f0} - D_f^{\mathrm{chance}}},\, 0,\, 1\right)
\end{equation}
and a retention score $r = \mathrm{clip}(D_t / D_{t0},\, 0,\, 1)$, averaged over the five seeds, where $D_{f0}$ and $D_{t0}$ are the pre-unlearning zero-shot forget and test accuracies. Arms whose zero-shot forget margin lies within five points of chance are excluded as degenerate. Selection is performed independently for each backbone--dataset pair, so the rate chosen for one dataset does not depend on the others.

\paragraph{Protocol.}
We evaluate on Flowers102, OxfordPets and Food101, unlearning one randomly drawn class per seed over five seeds per backbone--dataset pair. No augmentation is used: images are resized to $224$ and center cropped to $224\times224$ with bicubic interpolation, identically for unlearning and evaluation, and normalized with the standard CLIP statistics, mean $(0.4815, 0.4578, 0.4082)$ and standard deviation $(0.2686, 0.2613, 0.2758)$. Text context length is $77$. Evaluation builds a zero-shot classifier over all $C$ classes at a fixed logit scale of $100$; $D_f$ is accuracy on the full pool of held-out images of the forget class and $D_t$ is accuracy on the dataset test split, both reported before and after unlearning. Runs use FP32 for training and FP16 mixed-precision for evaluation, one GPU per run, with cuDNN in deterministic mode and benchmarking disabled.

\end{document}